\def\year{2022}\relax
\documentclass[letterpaper]{article} 

\usepackage{times}  
\usepackage{helvet}  
\usepackage{courier}  
\usepackage[hyphens]{url}  
\usepackage{graphicx} 
\usepackage{natbib}  
\usepackage{caption} 
\DeclareCaptionStyle{ruled}{labelfont=normalfont,labelsep=colon,strut=off} 
\usepackage{microtype}
\usepackage{etoolbox}

\usepackage{amsmath,eqnarray,mathtools}
\usepackage{amsfonts, dsfont, bm, courier}
\usepackage{graphicx,xcolor,placeins,caption}
\usepackage{enumitem}
\usepackage{subcaption,afterpage}
\newtheorem{definition}{Definition}

\providetoggle{submission}
\settoggle{submission}{false}
\providetoggle{aaai}
\settoggle{aaai}{true}
\providetoggle{neurips}
\settoggle{neurips}{false}

\usepackage{aaai22}  
\usepackage[utf8]{inputenc} 
\usepackage{nicefrac}       
\usepackage{microtype}      

\iftoggle{neurips}{
\usepackage[final]{neurips_2022}  
\setcitestyle{numbers,comma,square,sort}
\usepackage[font={footnotesize},labelfont={bf}]{caption}

}

\iftoggle{submission}{
\title{Predictive Multiplicity in Probabilistic Classification}
\author{
    Anonymous Author(s)
}
\affiliations{
    Affiliation\\
    email
}
}{
\iftoggle{aaai}{
\title{Predictive Multiplicity in Probabilistic Classification}
\author{
    Jamelle Watson-Daniels\textsuperscript{\rm 1}, David C. Parkes\textsuperscript{\rm 1,2}, Berk Ustun\textsuperscript{\rm 3}\\
}
\affiliations{
    \textsuperscript{\rm 1}Harvard University\\ \textsuperscript{\rm 2}DeepMind\\ \textsuperscript{\rm 3}U.C.~San Diego\\

    jwatsondaniels@g.harvard.edu, parkes@eecs.harvard.edu, berk@ucsd.edu
} }
}

\usepackage{array,booktabs,tabularx,multirow,colortbl,hhline}
\newcommand{\cell}[2]{\setlength{\tabcolsep}{0pt}\begin{tabular}{#1}#2 \end{tabular}}

\newcolumntype{H}{>{\setbox0=\hbox\bgroup}c<{\egroup}@{}}

\usepackage{tikz}
\newcommand*\circled[1]{\tikz[baseline=(char.base)]{
            \node[shape=circle,draw,inner sep=2pt] (char) {#1};}}

\setitemize{leftmargin=1em,parsep=2pt,label=\raisebox{0.25ex}{\tiny$\bullet$}}
\setlist[enumerate]{leftmargin=*, label= {\arabic*.}, itemsep=0.5em}
\newlist{thmlist}{enumerate}{1}
\setlist[thmlist]{leftmargin=*,label=\raisebox{0.25ex}{\tiny$\bullet$}, topsep=0.2em,itemsep=2pt}

\newcommand{\textds}[1]{{\footnotesize\texttt{#1}}}

\newcommand{\indic}[1]{\mathds{1}[#1]}

\newcommand{\dotprod}[2]{\langle{#1},{#2} \rangle}

\newcommand{\txplus}[0]{+}
\newcommand{\txminus}[0]{-}

\newcommand{\R}{\mathbb{R}}

\newcommand{\data}{\mathcal{D}}
\newcommand{\X}{\mathcal{X}}
\newcommand{\Y}{\mathcal{Y}}

\newcommand{\nplus}[0]{n^{\txplus{}}}
\newcommand{\nminus}[0]{n^{\txminus{}}}

\newcommand{\lossfun}[1]{L(#1)}

\newcommand{\wb}{\bm{w}}
\newcommand{\xb}{\bm{x}}

\newcommand{\st}{\textnormal{s.t.}}
\newcommand{\miprange}[3]{{#1}={#2},...,{#3}}

\newcommand{\clf}[0]{g}

\newcommand{\baseclf}[0]{\clf{}_0}

\newcommand{\Hset}[0]{\mathcal{H}}
\newcommand{\epsset}[1]{\Hset{}_\epsilon({#1})}

\newcommand{\predint}[2]{V_{{#1}}({#2})}
\newcommand{\ambiguity}[4]{A_{{#1},{#2}}({#3};{#4})}
\newcommand{\discrepancy}[4]{D_{{#1},{#2}}({#3};{#4})}

\newcommand{\upperbound}[0]{U_{i, \delta} }
\newcommand{\lowerbound}[0]{B_{i, \delta}}

\newcommand{\Wmax}[0]{W^\textrm{max}}
\newcommand{\Wmin}[0]{W^\textrm{min}}
\newcommand{\MLower}[0]{M_{v, i}}
\newcommand{\MUpper}[0]{M_{z,i}}
\newcommand{\binaryVarUpper}[0]{z_{i, \delta}}
\newcommand{\binaryVarLower}[0]{v_{i, \delta}}

\newcommand{\scr}[0]{s_{w}}

\begin{document}
\frenchspacing
\abovecaptionskip=2pt
\belowcaptionskip=0pt
\abovedisplayskip=2pt
\belowdisplayskip=2pt
\arraycolsep=0pt
\floatsep=0pt
\setlength{\tabcolsep}{2pt} 
\title{Predictive Multiplicity in Probabilistic Classification}

\maketitle
\begin{abstract} 
Machine learning models are often used to inform real world risk assessment tasks: predicting consumer default risk, predicting whether a person suffers from a serious illness, or predicting a person's risk to appear in court. Given multiple models that perform almost equally well for a  prediction task, to what extent do predictions vary across these models? If predictions are relatively consistent for similar models, then the standard approach of choosing the model that optimizes a penalized loss suffices. But what if predictions vary significantly for similar models? In machine learning, this is referred to as \emph{predictive multiplicity} i.e. the prevalence of conflicting predictions assigned by near-optimal competing models. In this paper, we present a framework for measuring predictive multiplicity in probabilistic classification (predicting the probability of a positive outcome). We introduce measures that capture the variation in risk estimates over the set of competing models, and develop optimization-based methods to compute these measures efficiently and reliably for convex empirical risk minimization problems. We demonstrate the incidence and prevalence of predictive multiplicity in real-world tasks. Further, we provide insight into how predictive multiplicity arises by analyzing the relationship between predictive multiplicity and data set characteristics (outliers, separability, and  majority-minority structure). 
 Our results emphasize the need to report predictive multiplicity more widely.
\end{abstract}
\section{Introduction}
\label{Sec::Introduction}
Probabilistic classification is often incorporated into real-world risk assessment tasks to inform decisions. For instance, probabilistic classifiers that predict consumer default risk are used by lenders to underwrite loans~\citep[][]{Bekhet2014,Attigeri2017}. Similarly in clinical applications, physicians make treatment decisions using models that predict whether a person suffers from a serious illness~\citep[][]{Than2014DevelopmentProtocol,Khand2017HeartValue,chen2021probabilistic}. In criminal justice, judges often make parole and sentencing decisions guided by models that predict the probability that a person will fail to appear in court~\citep[][]{Austin2010KentuckyValidation,Latessa2010TheORAS,christin2015courts,zeng2017interpretable}. 

The standard approach to selecting a probabilistic classifier often involves optimizing a loss function via empirical risk minimization. But for a given prediction task, there may exist multiple models that perform almost equally well, which is referred to in machine learning as model \emph{multiplicity} \citep[][]{Breiman2001}. These near-optimal, \emph{competing models}, have similar performance but characteristic differences - e.g. their interpretability~\citep[][]{Semenova2019a}, explainability~\citep[][]{Fisher2019,Dong2019}, counterfactual invariance~\citep[][]{DAmour2020}, or fairness~\citep[][]{Coston2021,Black2021LOOunffairness,Ali2021}. These differences can drastically change how we develop, choose, and use models~\citep{black2022model}.

We investigate how predictions change across competing models by studying \emph{predictive multiplicity}: the prevalence of conflicting predictions over competing models~\citep[][]{Marx2019}. To understand our motivation, consider the significance of competing models assigning vastly different predictions in practice.
In mortality prediction, a conflicting risk prediction would alter treatment decisions and health outcomes~\citep[][]{Moreno2005SAPSAdmission}. In drug discovery, a conflicting risk prediction could switch the compounds chosen for confirmatory experiments~\citep[][]{stokes2020deep}. By measuring and reporting the prevalence of conflicts, we can improve how we choose and use machine learning models. If end-users know that an individual risk estimate conflicts over the set of competing models, they could abstain from prediction~\citep[][]{Black2021,hamid2017machine} or defer a decision to a human expert~\citep[][]{mozannar2020consistent,kompa2021second}. If model developers know that many risk estimates conflict when compared across competing models, they might reconsider deployment and dedicate time to contend with multiplicity.
These implications underline the importance of measuring and reporting predictive multiplicity more widely.

%
%
Our main contributions are:
\begin{enumerate}
    \item We introduce measures of predictive multiplicity in our setting. The Viable Prediction Range examines how multiplicity affects predictions. Ambiguity and discrepancy reflect the proportion of individuals assigned conflicting risk estimates by competing models.
    
    \item We develop optimization-based methods to compute our measures for convex empirical risk minimization problems. This includes employing mixed-integer non-linear programming and outer-approximation algorithms. Whereas previous work defines competing models over a single performance metric, our methods enable developers to examine additional near-optimal metrics.
    
    \item We offer insights into why predictive multiplicity arises via systematic experiments on synthetic data. We find that predictive multiplicity is more prevalent for examples that are both outliers and close to the discriminant boundary, for datasets that are less separable, and for minority groups when a dataset has a majority-minority structure.
    
    \item We present an empirical study on seven real-world risk assessment tasks. We show that probabilistic classification tasks can in fact admit competing models that assign substantially different risk estimates. Our results also demonstrate how multiplicity can disproportionately impact marginalized individuals.
    
\end{enumerate}

\paragraph{Related Work.}
Our work is positioned alongside research on {\em model multiplicity}. This effect has been referenced in the statistics literature. 
For example, \citet{Chatfield1995} calls for performing a sensitivity analysis over competing models, while \citet{Breiman2001} cites multiplicity as a reason to avoid explaining a single model to draw conclusions about the broader data-generating process. Recent advances in computation make multiplicity analysis possible, leading to a stream of research on how competing models differ~\citep{Fisher2019,Dong2019,Semenova2019a,DAmour2020,veitch2021counterfactual,Pawelczyk2020,Coston2021,Black2021LOOunffairness,Ali2021}

\begin{figure}[t]
    \includegraphics[width=0.97\linewidth]{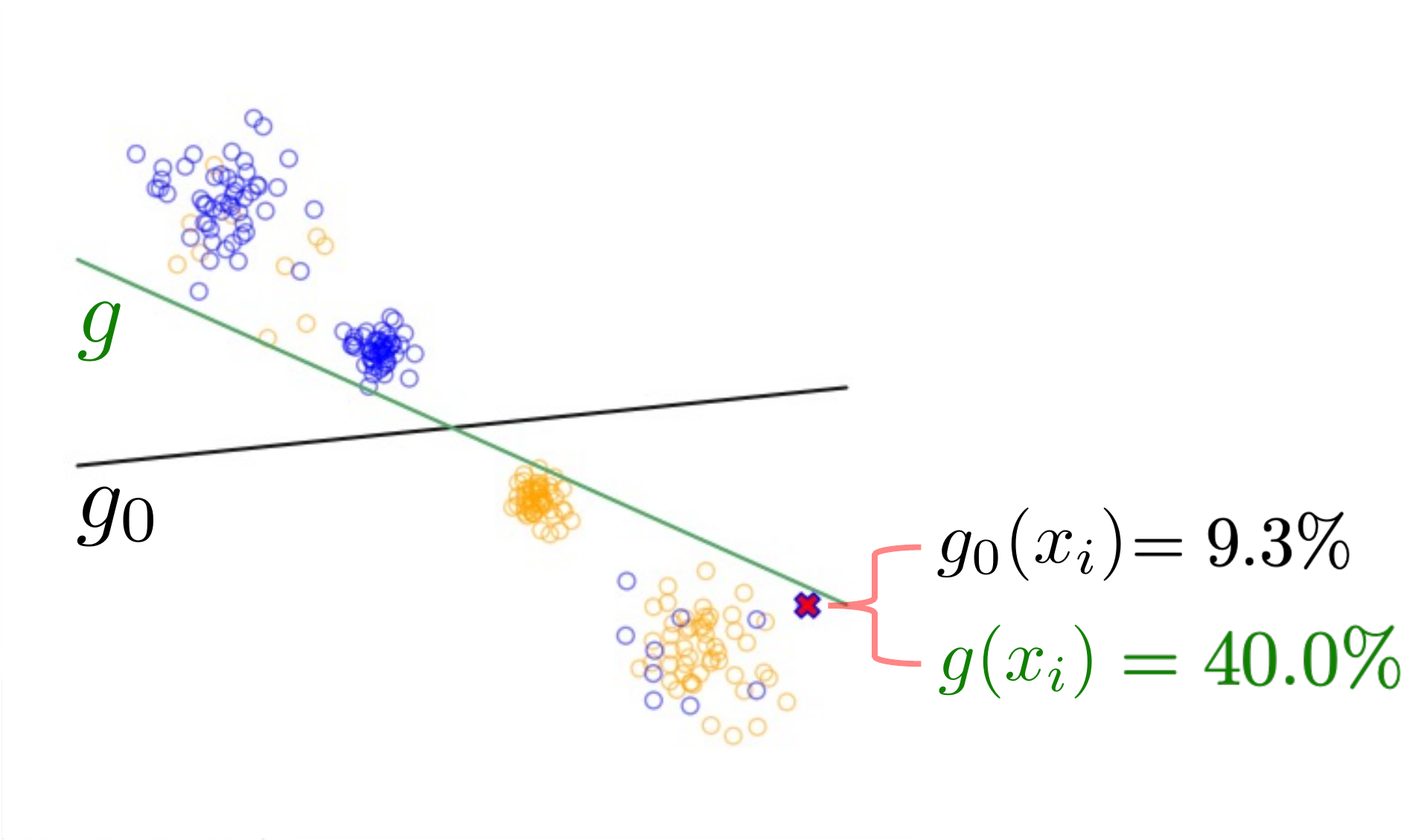}
    \caption{Classification models that make the same yes-or-no predictions can still assign conflicting risk predictions. Here, we show a 2D classification task with $\nplus{} = 200$ positive examples (blue) and $\nminus{} = 200$ negative examples (orange). We plot the decision boundary of a baseline model $\baseclf$ (black; log-loss/AUC/calibration = 0.41/0.88/17\%) and a competing model that performs almost equally well $\clf(\xb_i)$ (green; log-loss/AUC/calibration = 0.42/0.89/16\%). As shown, both classifiers make the same yes-or-no predictions, but assign conflicting risk estimates to individual examples e.g., example $\xb_i$ is assigned a risk estimate of $\baseclf(\xb_i) = 9.3 \%$ by the baseline model but $\clf(\xb_i) = 40.0\%$ by the competing model.}
    \label{Fig::ToyPredictionVsRiskAssessment} 
\end{figure}

Our work is distinctly focused on how multiplicity affects prediction. Our approach builds on \citet{Marx2019}, who study this effect in classification tasks with yes-or-no predictions. As shown in Figure~\ref{Fig::ToyPredictionVsRiskAssessment}, their measures and methods do not extend to our setting. 
Measuring multiplicity in probabilistic classification is complicated by the need to clarify the meaning of ``conflicting". In effect, what constitutes a conflicting risk prediction can change across applications (e.g., predictions that vary by 5\% or 30\%). Likewise, what constitutes a ``competing" model can change across applications. The present work addresses both of these problems by introducing methods that allow users to specify what is ``competing" (near-optimal metric) and what is ``conflicting" (deviation threshold). Also, previous work has yet to examine why predictive multiplicity arises, which we contribute to.

One way we compute predictive multiplicity is by constructing a range of individual risk predictions as a way to quantify pointwise uncertainty resulting from an underspecified empirical risk minimization problem. This relates to methods for evaluating predictive uncertainty such as conformal prediction~\citep[]{Shafer2008, Romano2020} as well as  Bayesian approaches~\citep[see e.g.,][]{Dusenberry2020,Lum2021}. However, conformal prediction focuses on uncertainty that arises due to non-conformity between historical data and new data, which is orthogonal to our goal. We focus on a non-Bayesian approach, recognizing that non-Bayesian methods are very typical in applied machine learning. 
Our goals relate also to a line of work that aims to quantify and communicate uncertainty in machine learning~\cite{Hofman2020, Kale2020, McGrath2020, Soyer2012, Kompa2021, wei2022on} and calibrate trust among stakeholders~\cite{Joslyn2013}. Other complementary work seeks interventions to resolve multiplicity~\citep[][]{Ali2021} or ensembling~\cite{Black2021}.
\section{Framework}
\label{Sec::Framework}

We consider a probabilistic classification task with a dataset of $n$ examples $\data{} = \{(\xb_i, y_i)\}_{i=1}^{n}$. Each example consists of a feature vector $\xb_i = [1,x_{i1},\ldots,x_{id}] \in \X \subseteq \R^{d+1}$ and a label $y_i \in \Y = \{-1, +1\}$, where $y_i =+1$ is an event of interest (e.g., default on a loan). With the dataset, we train a probabilistic classifier $\clf{}: \X \to [0,1]$ -- i.e., a model that assigns a risk estimate to example $\xb_i$ as: $\clf{}(\xb_i) := \text{Pr}(y_i = +1 | \xb_i) $. We refer to this model as the \emph{baseline model}, $\baseclf{}$, because it is the optimal solution to an empirical risk minimization (ERM) problem of the form:
\begin{align} 
\label{Opt::ERM}
 \min_{\clf{} \in \Hset} \lossfun{\clf{}; \data{}}, 
\end{align}
where
$\Hset$ is a family of probabilistic classifiers, and $L(~\cdot~; \data{})$ is a loss function evaluated on the dataset $\data{}$. In what follows, we write $\lossfun{\clf{}}$ instead of $\lossfun{\clf{}; \data{}}$ for conciseness. We evaluate the performance of a model in terms of $\lossfun{\clf{}}$, as well as the following metrics:
\begin{enumerate}[leftmargin=*]

\item \emph{Risk Calibration}: A risk-calibrated model assigns risk predictions that match observed frequencies~\citep{Naeini2015BinaryApproach}. We measure risk calibration in terms of {\em expected calibration error}:
\begin{align}
\label{Eq::ECE}
\textrm{ECE}(\clf{}) = \sum_{b=1}^B \frac{n_b}{n} |\hat{p}_b(\clf{}) - \bar{p}_b|.
\end{align}
Here: $I_b$ is the index set of $n_b$ examples in bin $b \in [B]$; and $\hat{p}_b(\clf{}) := \tfrac{1}{n_b}\sum_{i \in I_b}~\clf{}(\xb_i)$ and $\bar{p}_b = \tfrac{1}{n_b}\sum_{i \in I_b} \indic{y_i = +1}$ are the mean predicted risk and mean observed risk of examples in bin $b \in [B]$, respectively. 

\item \emph{Rank Accuracy}: A rank-accurate model outputs risk predictions that can be used to correctly order examples in terms of true risk. We assess  rank accuracy using the \emph{area under the ROC curve}:
\begin{align} 
\label{Eq:AUC}
\textrm{AUC}(\clf{}) = \frac{1}{\nplus \nminus} \sum_{\substack{i: y_i = +1\\ k: y_k = -1}} \indic{\clf{}(\xb_i) > \clf{}(\xb_k)},
\end{align}
where $\nplus{} = |\{i: y_i = +1\}|$ and $\nminus{} = |\{i: y_i = -1\}|$. 

\end{enumerate}

In what follows, we let $M(\clf{};\mathcal{D})\in \R_+$ denote the performance of $\clf{}\in \Hset$ over a dataset $\mathcal{D}$ in regards to {\em performance metric} $M(g)$, where the convention is that lower values of $M(g)$ are better; when working with AUC, we measure the \emph{AUC error}: $M(g) = 1 - \textrm{AUC}(g)$.

\subsection{Competing Models}

Competing models are classifiers with near-optimal performance compared to the baseline model. A \emph{competing model} is any model $g \in \Hset$ whose performance is within $\epsilon$ of the baseline model $\baseclf{}$.
\begin{definition}[$\epsilon$-Level Set]
Given a baseline model $\baseclf{}$, 
metric $M$, and error tolerance $\epsilon>0$, the \emph{set of competing models} ($\epsilon$-level set) is the set:
\begin{align*}
  \!  \epsset{\baseclf{}} \! :=\!  \{\clf{} \in \Hset: M(\clf{}) \leq M(\baseclf{})  +  \epsilon\}.
\end{align*}
\end{definition}

Our methods consider multiplicity over a range of  $\epsilon$ values. In practice, a suitable choice  of $\epsilon$ should  reflect the epistemic uncertainty in the performance of the baseline model. For instance, one could employ bootstrap re-sampling to measure the model uncertainty due to sample variation or consider worst-case uncertainty through  generalization bounds.  

\subsection{Measuring Viable Risk Predictions}

To examine how multiplicity affects predictions, we define a range of viable risk estimates that can be assigned by competing models. 
\begin{definition}[Viable Prediction Range]
The {\em  viable prediction range} is the smallest and largest risk estimate assigned to example $i$ over competing models in the $\epsilon$-level set:
\begin{align}
    \predint{\epsilon}{\xb_i} := [\min_{\clf{} \in \epsset{\baseclf{}}} \clf{}(\xb_i), \max_{\clf{} \in \epsset{\baseclf{}}} \clf{}(\xb_i)].\label{Eq::Interval}
\end{align} 
\end{definition}

For a prediction task, computing the viable prediction ranges over a sample illuminates the extent to which competing models assign  different risk estimates to individuals. Although we express the prediction range over an $\epsilon$-level set using $[\cdot,\cdot]$ interval notation, not all predictions between the min and the max may be attainable by a competing model.

\subsection{Measuring Predictive Multiplicity}

We say that a risk estimate is {\em conflicting} if it differs from the baseline risk estimate by at least some deviation threshold, $\delta \in (0,1)$.
The appropriate value of $\delta$ will depend on the application; i.e. a conflicting risk prediction in a clinical decision support task may differ from that which constitutes a conflicting risk prediction in recidivism prediction. 


%
Ambiguity and discrepancy reflect the proportion of examples in a sample $S$ assigned conflicting risk estimates by competing models. These definitions follow~\citet{Marx2019}, who give analogous definitions for the problem of multiplicity  with binary predictions (see Figure~\ref{Fig::ToyPredictionVsRiskAssessment} for an illustration of the difference between this problem and the multiplicity of risk estimates).

\begin{definition}[Ambiguity]
\label{Def::Ambiguity}
The {\em $(\epsilon,\delta)$-ambiguity} of a probabilistic classification task over a sample $S$ is the proportion of examples in $S$ whose baseline risk estimate changes by at least $\delta$ over the $\epsilon$-level set: 
\begin{align*}
  \ambiguity{\delta}{\epsilon}{\baseclf{}}{S} 
  &:= 
  \frac{1}{|S|} \sum_{i\in S} \indic{ \max_{\clf{} \in \epsset{\baseclf{}}} | \clf{(\xb_i)} - \baseclf{}(\xb_i) |  \geq \delta}.
\end{align*}
\end{definition}
Relative to the baseline model, ambiguity makes a statement about the proportion of individuals whose risk estimate is uncertain by at least $\delta$. High ambiguity means more uncertainty in risk predictions. Users may also consult the viable prediction range to guide decisions using the baseline model.

\begin{definition}[Discrepancy]
\label{Def:: Discrepancy}
The $(\epsilon,\delta)$-discrepancy of a probabilistic classification  task over a sample $S$ is the maximum proportion of examples in $S$ whose risk estimates could change by at least $\delta$ by switching the baseline model with a competing model in the $\epsilon$-level set: 
\begin{align*}
\discrepancy{\delta}{\epsilon}{\baseclf{}}{S} :=  \max_{\clf{} \in \epsset{\baseclf{}}} \frac{1}{|S|} \sum_{i\in S} \indic{ |\clf{}(\xb_i) - \baseclf{}(\xb_i)| \geq \delta }.
\end{align*} 
\end{definition}

Relative to the baseline model, discrepancy reflects the maximum the number of conflicting risk estimates as a result of replacing baseline model with a competing model in the $\epsilon$-level set. 

Ambiguity and discrepancy differ in the stance they take in regard to the worst case. Discrepancy measures the worst-case number of predictions that will change by switching the baseline model with a competing model. In contrast, ambiguity focuses on the worst case for prediction variation over the set of competing models. If we were to abstain from prediction on points that are assigned a conflicting prediction by a competing model~\citep[using e.g., selective classification methods][]{Black2021}, then ambiguity would reflect the abstention rate.

\begin{figure*}[t]
    \centering
    \includegraphics[width = 0.75\linewidth]{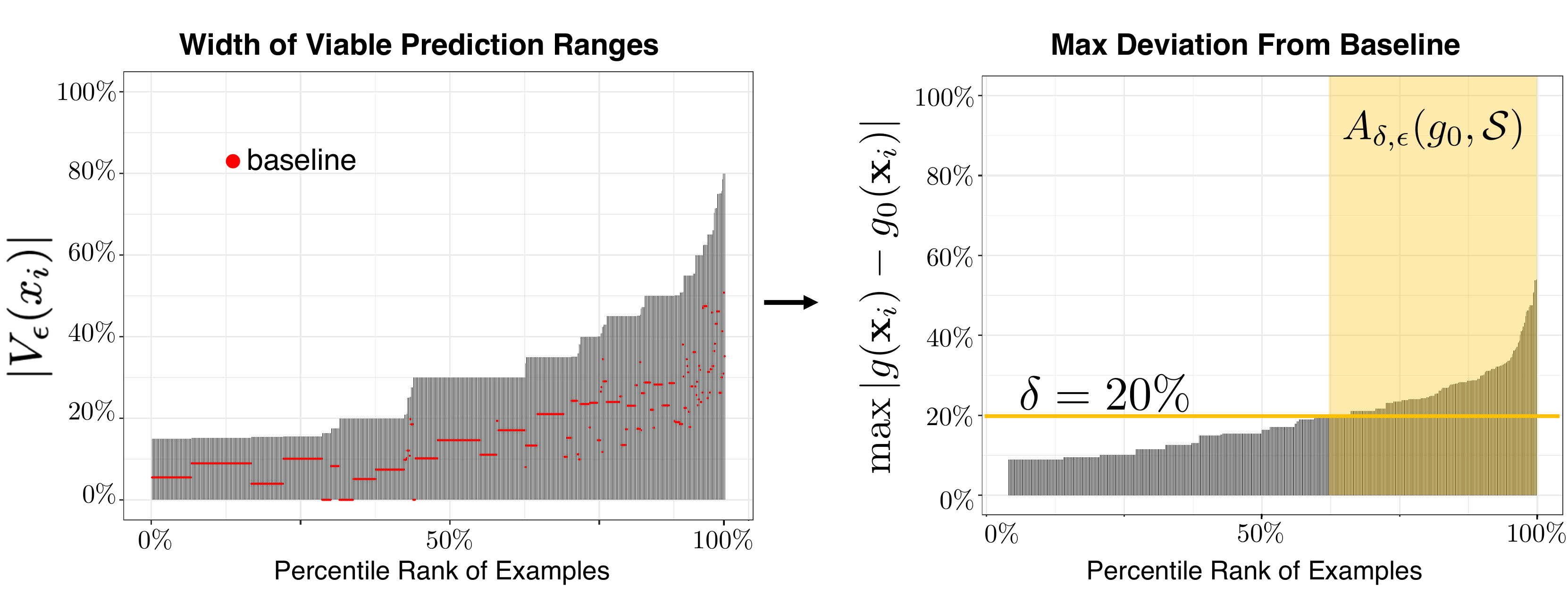}
    \caption{An illustration of how viable prediction ranges relate to ambiguity. Left, we plot the width of the viable prediction ranges $\left| \predint{\epsilon}{\xb_i} \right|$ on the $y$-axis for each example on the $x$-axis. Note that widths shifted to start from zero and examples shown in increasing order. The plot also shows the individual baseline risk estimates for each example in red $\baseclf{(\xb_i)} -  \min_{\clf \in \mathcal{H}_\epsilon (\baseclf)} \clf{}(\xb_i)  $ (shifted similarly). To interpret, the first example from the left has a width of $\approx 15\%$ with a baseline risk estimate on the lower side of the range. The last example has a width of $\approx 80\%$ with a baseline risk estimate closer to the higher side of the range. Using the viable prediction ranges $\predint{\epsilon}{\xb_i}$ directly, we can extract the maximum difference from the baseline. On the right, we plot the maximum deviation from the baseline, $\max \left| \clf{}(\xb_i) - \baseclf{}(\xb_i) \right|$ for each example on the $x$-axis (increasing order). To interpret, consider a deviation threshold $\delta = 20\%$, all examples with max deviation above that threshold are highlighted in yellow giving us ambiguity, $\ambiguity{\delta}{\epsilon}{\baseclf{}}{S}$. \label{fig:range_to_amb} }
\end{figure*}

\paragraph{Computing Ambiguity with Viable Prediction Ranges.}
As shown in Figure~\ref{fig:range_to_amb}, we can use the viable prediction ranges of all points in a sample to compute ambiguity. Given the viable prediction range for each example, we can calculate the maximum difference between the baseline risk and that assigned by competing models. We can then compute ambiguity by measuring the proportion of examples where this difference exceeds the deviation threshold.

\section{Methodology}
\label{Sec::Methodology}
In this section, we detail the procedure for computing measures of predictive multiplicity. This methodology can be applied to any convex loss function $L(\cdot)$, and together with a training problem that employs a convex regularization term.
We illustrate the methodology on the classification task described in \S \ref{Sec::Framework} by training a probabilistic classifier via logistic regression, with $\clf{}(\xb_i) = \frac{1}{1 + \exp(- \langle \wb, \xb_i \rangle )}$
, where  $\wb = [w_0, w_1,\ldots, w_d ]^\top \in \R^{d+1}$ is a coefficient vector. We train this baseline model by solving Eq.~\eqref{Opt::ERM} to minimize normalized {\em  logistic loss}: $\lossfun{\wb} = \tfrac{1}{n} \sum_{i = 1}^n  \log (1 + \exp(- \dotprod{\wb}{y_i\xb_i}))$. 
\newcommand{\pathset}[1]{\mathcal{G}_{#1}}
\subsection{Measuring Ambiguity}
\label{Sec::AmbiguityOpt}
We first present a method for computing ambiguity for different choices of $\epsilon$ and $\delta$. The method also gives a conservative approximation of the viable prediction range for each example.
We  construct a pool of {\em candidate models} that assign a specific risk estimate to each example. From these models, we select those with  performance within $\epsilon$ of the baseline model 
as the set of competing models. 
\begin{definition}[Candidate Model]
Given a baseline model $\baseclf{}$, a finite set of user-specified threshold probabilities $P \subseteq [0,1]$,
 then for each $p\in P$ a {\em candidate model} for example $\xb_i$ is an optimal solution to the following constrained ERM:
\begin{align}
\begin{split}
\min_{\wb \in \R^{d+1}} & \ \lossfun{\wb} \\
\st  \quad & \clf{}(\xb_i) \leq p, \;\; \textrm{if}~ p < \baseclf{}(\xb_i)\\
& \clf{}(\xb_i) \geq p.  \;\;\; \textrm{if} ~ p > \baseclf{}(\xb_i)
\end{split}\label{Opt::PathCLF}
\end{align}
\end{definition}

For each threshold probability $p\in P$, we train a candidate model $\clf{}$ such that the probability assigned to the example is constrained to the threshold $p$.
In this way, by training for each example and threshold probability $p\in P$,
we obtain the set of candidate models $\pathset{} :=\{ \clf{}: i \in S, p\in P\}$. 
 We choose to solve the instances in order of increasing values of  threshold probability $p$, which allows us to warm-start the optimization using previous solutions. 

Given the set of candidate models,  we define a {\em candidate $\epsilon$-level set} as
\begin{align}
\tilde{\mathcal{H}}_{\epsilon}(\baseclf{})&\!=\!\{g\in \pathset{} : M(\clf{})\!\leq\! M(g_0)\!+\!\epsilon\}.\label{Eq::PathEps}
\end{align}

We use the candidate $\epsilon$-level set to compute measures of  predictive multiplicity. This method is exact for ambiguity defined in terms of near-optimal loss when the grid of threshold probabilities $P \subseteq [0,1]$ aligns with $\baseclf{(x_i)} \pm \delta$ (i.e., is selected as appropriate to the baseline prediction for an example and the value of $\delta$).
For other metrics, such as AUC, this approach to compute ambiguity gives a conservative estimate (i.e., lower bound)---the training of a candidate model  does not directly optimize for AUC, but we can retain only those candidate models that are competitive for the appropriate $\epsilon$-level set definition. Since $\tilde{\mathcal{H}}_{\epsilon}(\baseclf{})\subseteq \epsset{\baseclf{}}$, the candidate-model approach also provides a  conservative estimate of the viable prediction range (Eq.~\eqref{Eq::Interval}) for an example.

\subsection{Measuring Discrepancy}
\label{Sec::MethodDiscrepancy}

Discrepancy  is the maximum proportion of examples assigned conflicting risk estimates by a single competing model, $\clf{} \in \epsset{\baseclf{}}$. Recall that a conflicting risk estimate differs from the baseline risk estimate $\baseclf{(\xb_i)}$ by at least some deviation threshold, $\delta > 0$. Therefore, measuring discrepancy with respect to a baseline model corresponds to solving the following maximization problem:
\begin{align}
\begin{split}\label{Eq::DiscERM}
    \max_{\clf{} \in \epsset{\baseclf{}}} &\qquad \sum_{i\in S}  \indic{| \clf{}(\xb_i) - \baseclf{(\xb_i)} | \geq \delta}.
\end{split}
\end{align}

Given a sample $S$, the baseline loss $L_0$, error tolerance $\epsilon$, and deviation threshold $\delta$, we can formulate Eq.~\eqref{Eq::DiscERM} as a mixed-integer non-linear program (MINLP):
\begin{subequations}
\label{Opt::DiscrepancyMINLP}
\begin{equationarray}{@{}c@{}r@{\,}c@{\,}l>{\,}l>{\,}r@{\;}}
    \max_{\wb \in \R^{d+1}} & \quad \sum_{i\in S} d_i & & & & \notag \\[0.75em] 
    \st 
    & \lossfun{\wb} & \leq & L_0 + \epsilon &  \label{Con::LossConstraint} \\
    & d_i & = & \binaryVarLower + \binaryVarUpper & \forall{i}\in S \ \ \  \label{Con::BigMUpper} \\ %
    & \MUpper (1 - \binaryVarUpper) &\geq & \dotprod{\wb}{\xb_i} - \upperbound{}  & \forall{i}\in S  \label{Con::BigMUpperOnlyIf} \\
    & \MLower (1 - \binaryVarLower) &\geq &-\dotprod{\wb}{\xb_i} + \lowerbound{} & \forall{i}\in S \label{Con::BigMLowerOnlyIf} \\
    & d_i, \binaryVarUpper, \binaryVarLower  & \in & \{0,1\} & \forall{i}\in S \nonumber 
\end{equationarray}
\end{subequations}

The MINLP in \eqref{Opt::DiscrepancyMINLP} fits the parameters of a linear classifier that maximizes discrepancy . 
Here, the objective maximizes number of examples assigned a conflicting risk estimate using the indicator variables $d_i := \indic{| \clf{}(\xb_i) - \baseclf{(\xb_i)} | \geq \delta}$. 
Each $d_i$ is set to $\binaryVarUpper:= \indic{\clf{}(\xb_i) \leq  ({\baseclf(\xb_i) -\delta})}$ (or $\binaryVarLower:= \indic{\clf{}(\xb_i) \geq  ({\baseclf(\xb_i) +\delta})}$) when the model assigns a risk estimate to example $i$ that exceeds $\delta$ on the low-side (or high-side) of the baseline risk estimate, respectively. 
We ensure the indicator behavior of $\binaryVarUpper$ and $\binaryVarLower$ through the ``Big-M" constraints~\eqref{Con::BigMLowerOnlyIf} and~\eqref{Con::BigMUpperOnlyIf}, which flag deviations in score space. The Big-M parameters can be set as $\MUpper := -\upperbound + \max_{\wb} \dotprod{\wb}{\xb_i}$ and $ \MLower :=  \lowerbound - \min_{\wb} \dotprod{\wb}{\xb_i}$, where $\upperbound := \textrm{logit}({\baseclf(\xb_i) -\delta})$, and $\lowerbound := \textrm{logit}({\baseclf(\xb_i) + \delta})$. 
When the values of $\upperbound$ and  $\lowerbound$ lie outside of the $[0,1]$ domain of the logit, we can drop the relevant indicator variable from the formulation. We provide additional details in the Appendix.
\paragraph{Outer-Approximation Algorithm.}
The challenge in solving~\eqref{Opt::DiscrepancyMINLP} is that constraint \eqref{Con::LossConstraint} is non-linear. 
We construct a linear approximation of the loss~\citep[see e.g.,][]{franc2008optimized,joachims2009cutting} using an iterative,  outer-approximation method~\citep[see e.g.,][]{ustun2016kdd,bertsimas2016best,bertsimas2017logistic} to solve.
The algorithm recovers a globally optimal solution to the MINLP in~\eqref{Opt::DiscrepancyMINLP}, and can be implemented using a mixed-integer programming solver with callback functions~\citep[see e.g.,][]{ustun2016kdd,bertsimas2016best,bertsimas2017logistic}. 
The procedure builds a branch-and-bound tree to discover integer-feasible solutions that obey all constraints other than \eqref{Con::LossConstraint}. 
For each feasible solution identified, the procedure computes its loss to determine if it is feasible with respect to constraint \eqref{Con::LossConstraint}. If feasible, the procedure retains the solution. Otherwise, it updates the loss function approximation by adding a new linear constraint. 

This method is exact for computing discrepancy in terms of near-optimal loss. For other metrics, we can again treat the intermediate solutions to the outer-approximation algorithm as candidate models and use these candidates to recover a lower bound similar to the method used in \S~\ref{Sec::AmbiguityOpt}.

\newcommand{\fixedwidthcell}[2]{{\setlength{\tabcolsep}{0pt}\begin{tabular}{p{#1}}{#2}\end{tabular}}}
\newcommand{\addtp}[1]{%
\cell{c}{{\scriptsize\includegraphics[trim=9.0em 5.5em 20em 7.0em, clip, width=0.2\linewidth]{{#1}}}}%
}

\newcommand{\addicmltp}[1]{%
\cell{c}{{\scriptsize\includegraphics[trim=9.0em 5.7em 20em 7.2em, clip, width=0.49\linewidth]{{#1}}}}%
}

\newcommand{\addsmalltp}[1]{%
\cell{c}{{\scriptsize\includegraphics[trim=9.0em 5.7em 20em 7.2em, clip, scale = 0.35]{{#1}}}}%
}

\newcommand{\addarxivtp}[1]{%
\cell{c}{{\scriptsize\includegraphics[trim=10.0em 6.0em 20em 7.3em, clip, scale = 0.25]{{#1}}}}%
}

\section{Numerical Experiments}
\label{Sec::Experiments}

In this section, we present experiments on synthetic and real-world data. Our goals are to: (1) reveal dataset characteristics that impact predictive multiplicity; and (2) determine the extent to which real risk assessment tasks exhibit predictive multiplicity in practice.

\subsection{Synthetic Datasets}
\label{Sec::ToyExamples}

\paragraph{Linear Separability.}
To demonstrate how separability informs predictive multiplicity, we compute ambiguity while varying the degree of separability and show results in Figure~\ref{Fig::ToyExamples} column \textbf{(A)}. We set $\delta= 20\%$ and $\epsilon = 5\%$ and control separability by increasing the variance of the data from $\sigma = 4$ (top) to $\sigma = 10$ (bottom). A clear trend is that ambiguity increases as the data becomes less separable from $1\%$ to $21\%$. Notice, also that the ambiguous examples tend to be those near the discriminant boundary and outliers. 

\paragraph{Outliers and Margin Distance.}
We examine how predictive multiplicity relates to outlier distance from the discriminant boundary. We position outliers near and far from the discriminant boundary and compute ambiguity. As shown in Figure~\ref{Fig::ToyExamples} column \textbf{(B)}, a clear trend is that examples that are outliers but far from the discriminant boundary (high margin) are less susceptible to predictive multiplicity. 

\paragraph{Majority-Minority Structure.}
We consider the effect of systematically varying the majority-minority structure of data. For this, we generate a majority class that has a different statistical pattern of features than a minority class. Given the two groups, the model is faced with a tradeoff between correctly predicting one group or the other. In Figure~\ref{Fig::ToyExamples} column \textbf{(C)}, we vary the ratio in a majority-minority structure revealing that the minority group is more prone to predictive multiplicity. The ambiguity of the minority group at 10:1 is substantially larger than for the majority group. This shows the importance of evaluating multiplicity across subgroups.
\newcommand{\addtpNOcrop}[1]{%
\cell{c}{{\scriptsize\includegraphics[scale = 0.3]{{#1}}}}%
}

\begin{figure*}[htb]
     \centering
     \scriptsize
     \resizebox{0.8\linewidth}{!}{\begin{tabular}{c|c|c}
     \circled{A} & \circled{B} & \circled{C} \\[0.3em] 
     More Separable, $\ambiguity{\delta}{\epsilon}{\baseclf{}}{S} = 1\%$ & 
     Large Margin, $\ambiguity{\delta}{\epsilon}{\baseclf{}}{S} = 1\%$ & 
     Ratio 10:1,  $\ambiguity{\delta}{\epsilon}{\baseclf{}}{S} = 7\%$\\
     \addtpNOcrop{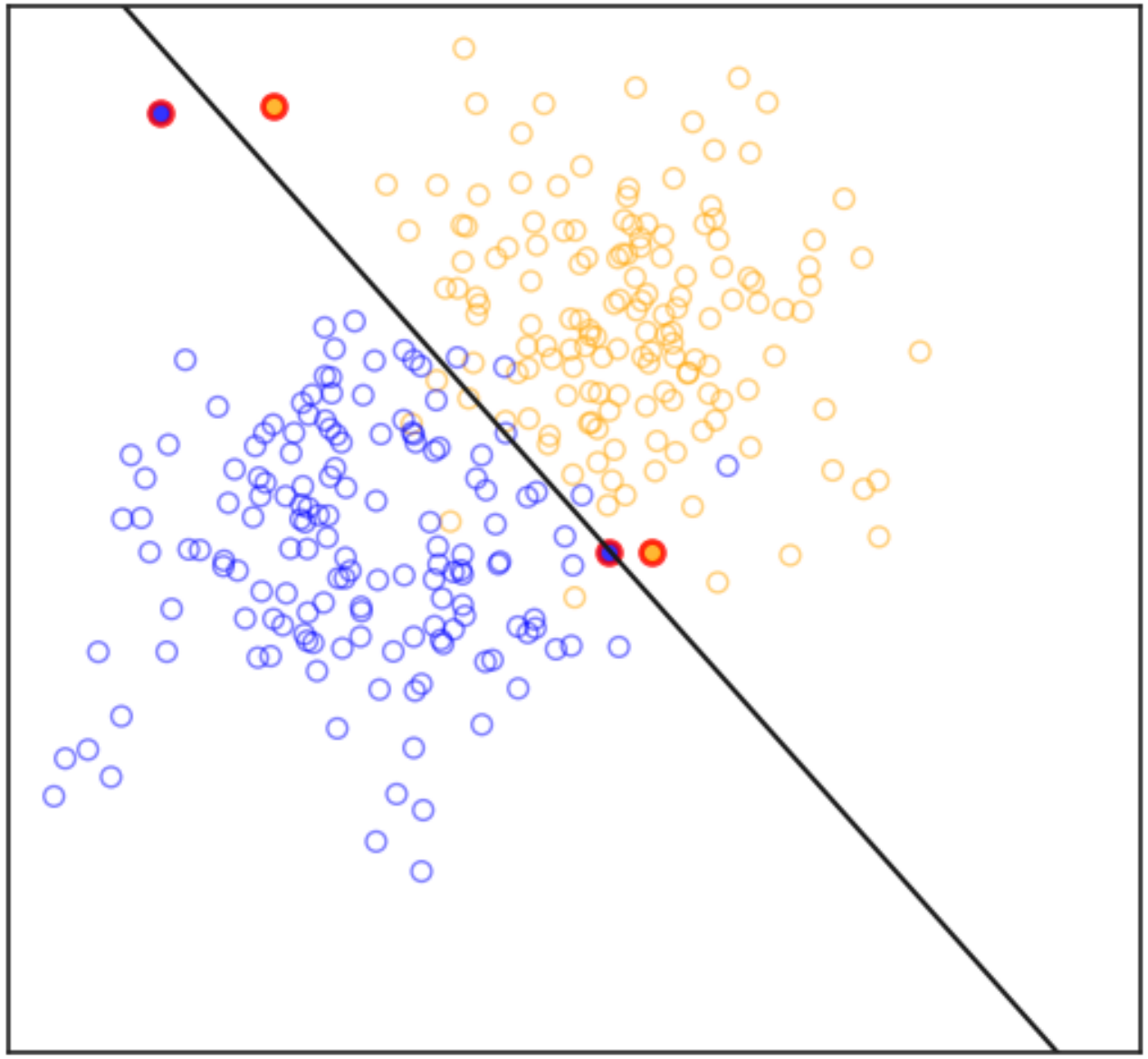} &
     \addtpNOcrop{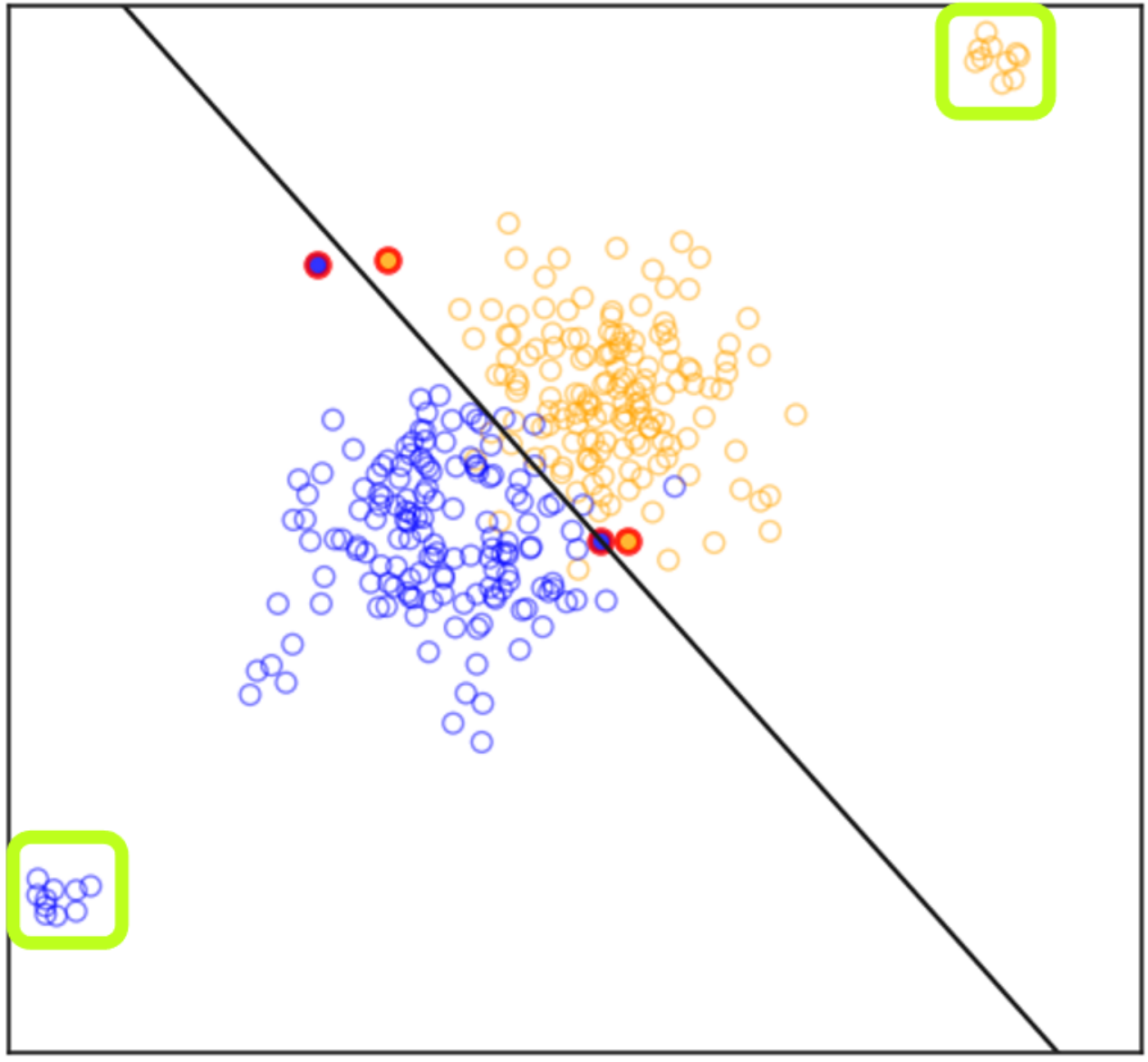} &
     \addtpNOcrop{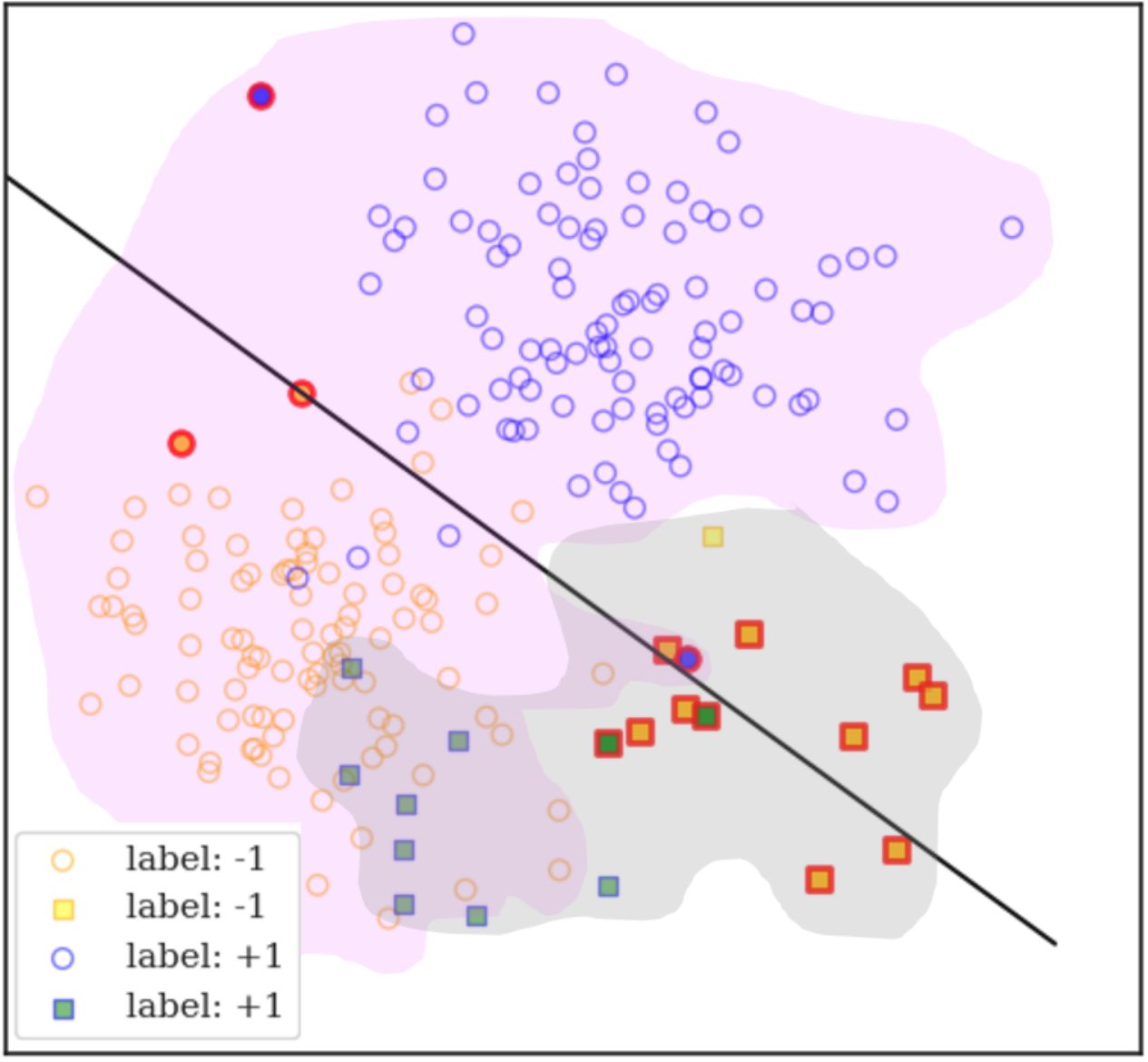} \\
    Less Separable, $\ambiguity{\delta}{\epsilon}{\baseclf{}}{S} = 21\%$ & 
    Small Margin, $\ambiguity{\delta}{\epsilon}{\baseclf{}}{S} = 3\%$ & 
    Ratio 1:1, $\ambiguity{\delta}{\epsilon}{\baseclf{}}{S} = 53\%$ \\
    \addtpNOcrop{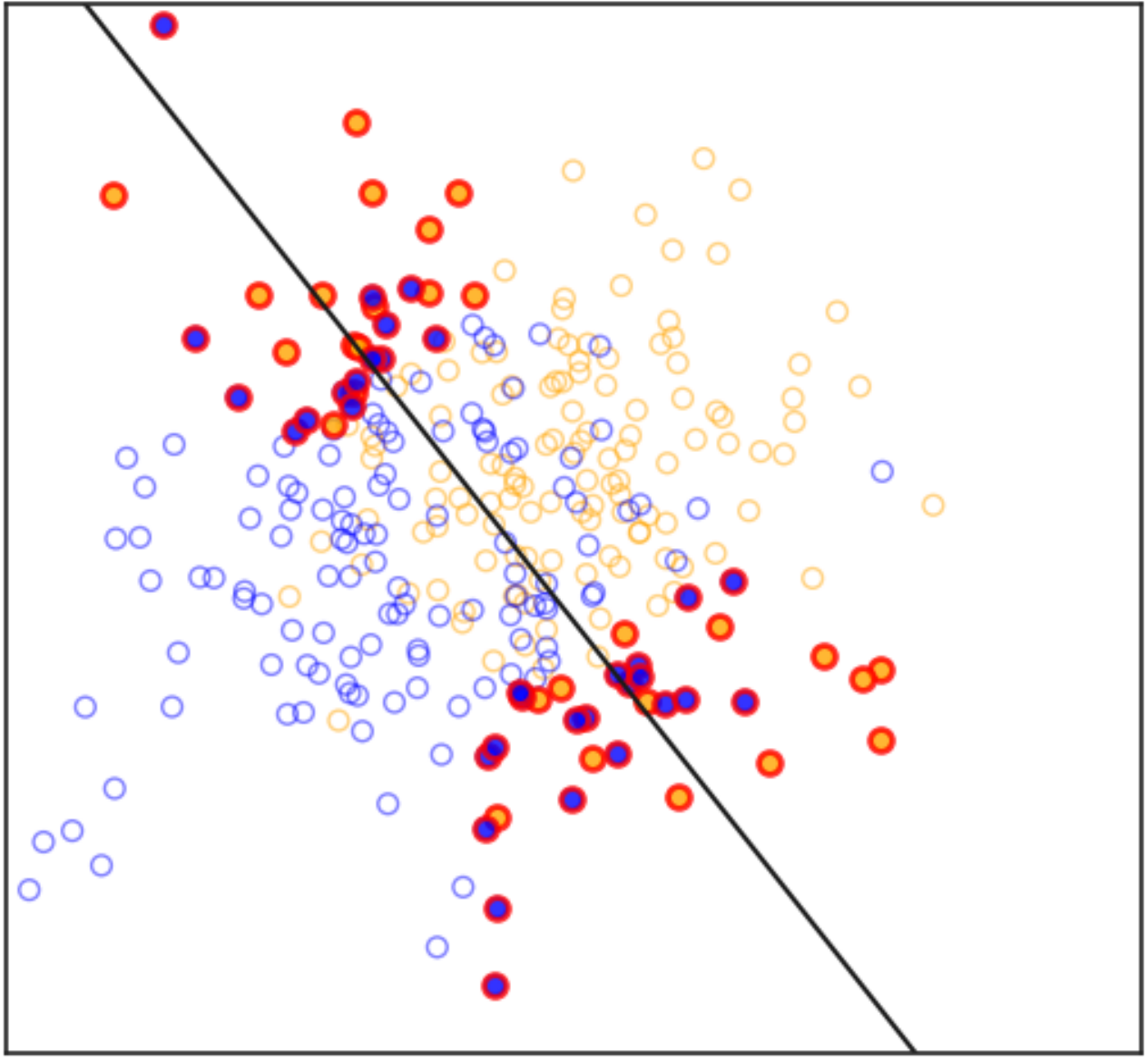} &
     \addtpNOcrop{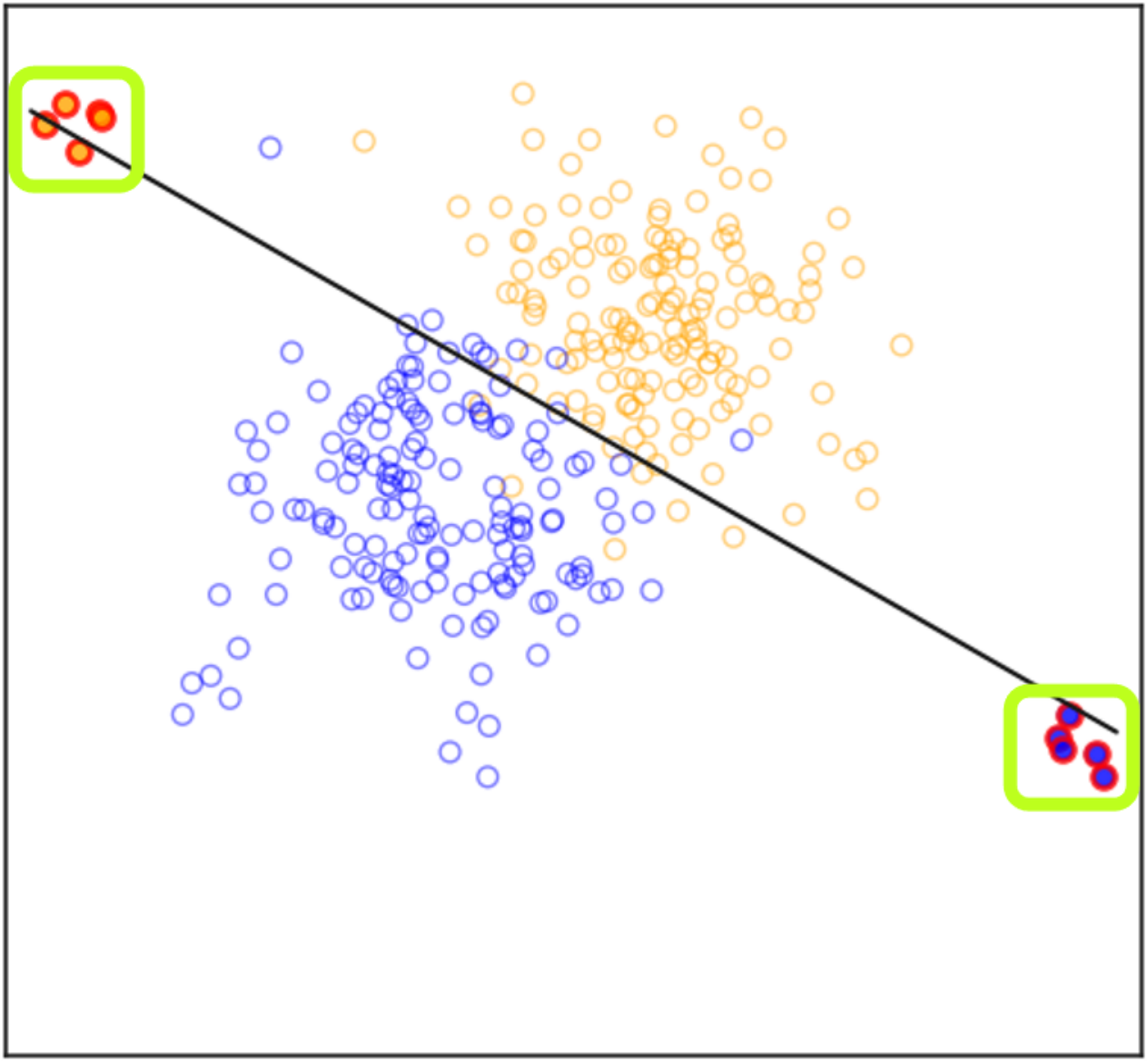} &
     \addtpNOcrop{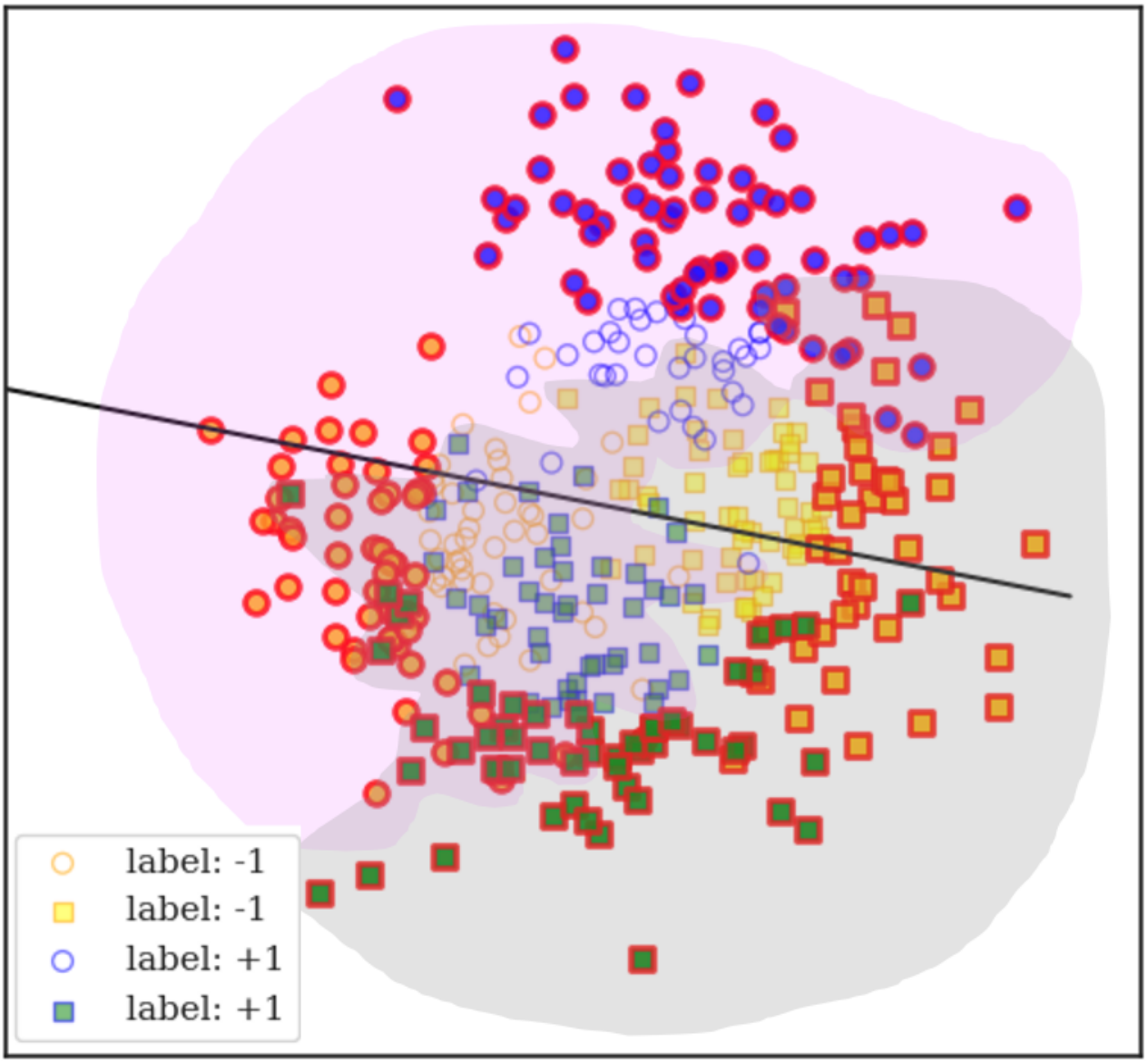}
     \end{tabular}}
      \caption{Experiments on synthetic data. In (A), we vary separability and find that ambiguity increases as separability decreases. In (B), we position outliers near and away from the discriminant boundary finding that outliers closer to the boundary are more prone to ambiguity. In (C), we vary the ratio in a majority-minority structured dataset: magenta shading- majority group (circles), grey shading- minority group (squares) revealing that the minority group is more prone to ambiguity. In the figures, $Y=+1$ examples are blue, $Y=-1$ examples are orange, and ambiguous examples are highlighted red and we set $\delta= 20\%$ and\label{Fig::ToyExamples} $\epsilon = 5\%$.}
\end{figure*}


\newcommand{\adddata}[2]{{\cell{l}{\texttt{\scriptsize{#1}}} \\ \scriptsize{#2}}}
\newcommand{\adddumbbell}[1]{%
\cell{c}{\includegraphics[width=0.2\linewidth, trim=0.2cm 0.5cm 0.0cm 0.75cm,clip]{{#1}}}%
}

\newcommand{\addhm}[1]{%
\cell{c}{\includegraphics[width=0.25\linewidth,trim=0cm 0.0cm 2.5cm 1.0cm,clip]{{#1}}}%
}

\newcommand{\addhmNOCROP}[1]{%
\cell{c}{\includegraphics[width=0.25\linewidth]{{#1}}}%
}

\newcommand{\adddisc}[1]{%
\cell{c}{\includegraphics[width=0.195\linewidth,trim=0.75cm 0.0cm 2.5cm 1.0cm,clip]{{#1}}}%
}

\newcommand{\hmdeltalabels}[1]{%
\includegraphics[width=0.02\linewidth,trim=0 0 15.7cm 0,clip]{{#1}}%
}

\newcommand{\addinterval}[1]{%
\cell{c}{\includegraphics[width=0.29\linewidth]{{#1}}}%
}

\newcommand{\addhmFINAL}[1]{%
\cell{c}{\includegraphics[width=0.25\linewidth]{{#1}}}%
}

\begin{figure*}[t!]
    \centering
    \resizebox{0.78\linewidth}{!}{\begin{tabular}{m{0.5cm}ccc} & 
    \texttt{\normalsize{mammo:}} \normalsize{breast cancer} &
    \texttt{\normalsize{apnea:}} \normalsize{sleep apnea} &
    \texttt{\normalsize{arrest:}} \normalsize{crime rearrest} \\
    \toprule
    {\circled{A} } &
    \addinterval{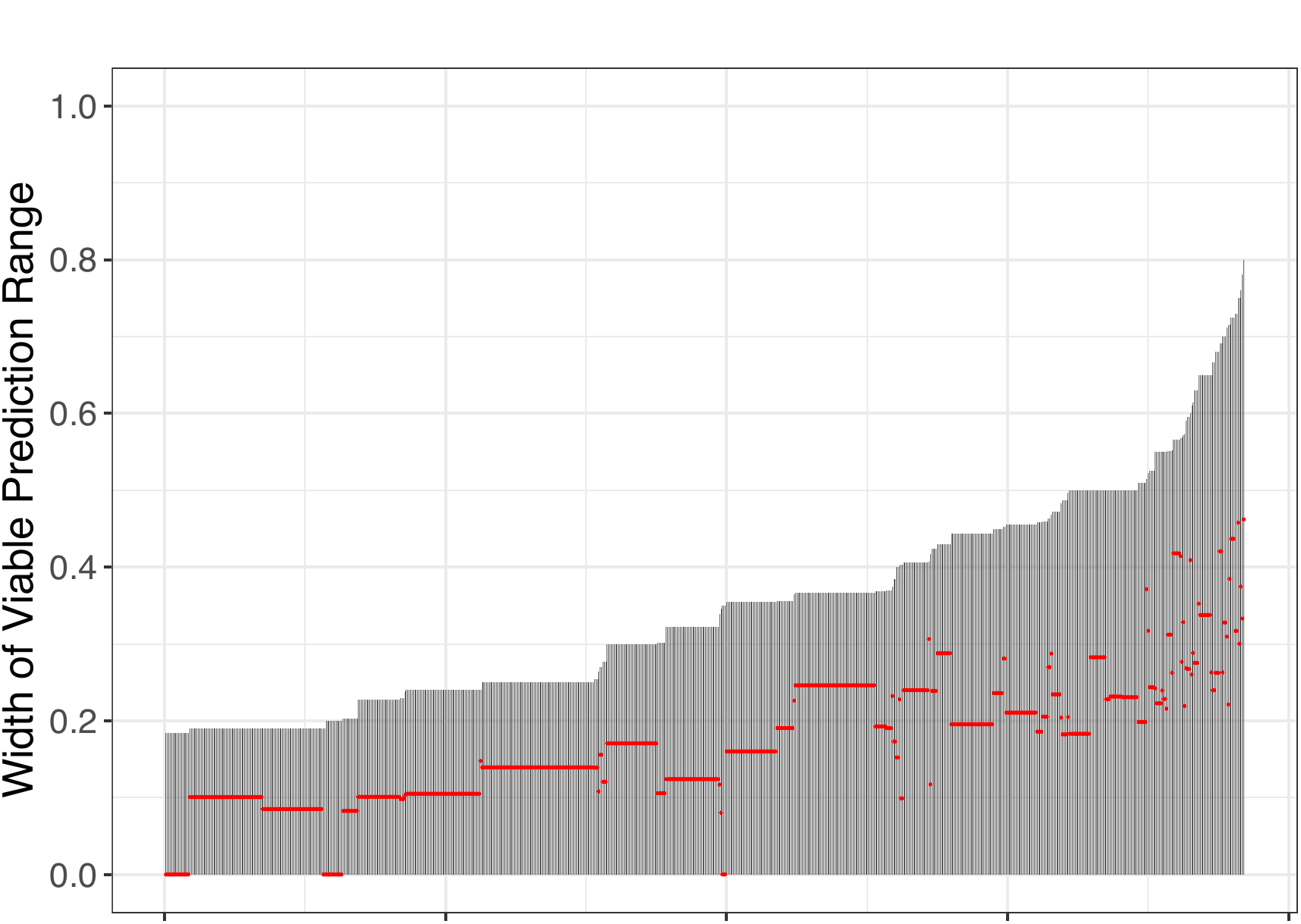} &
    \addinterval{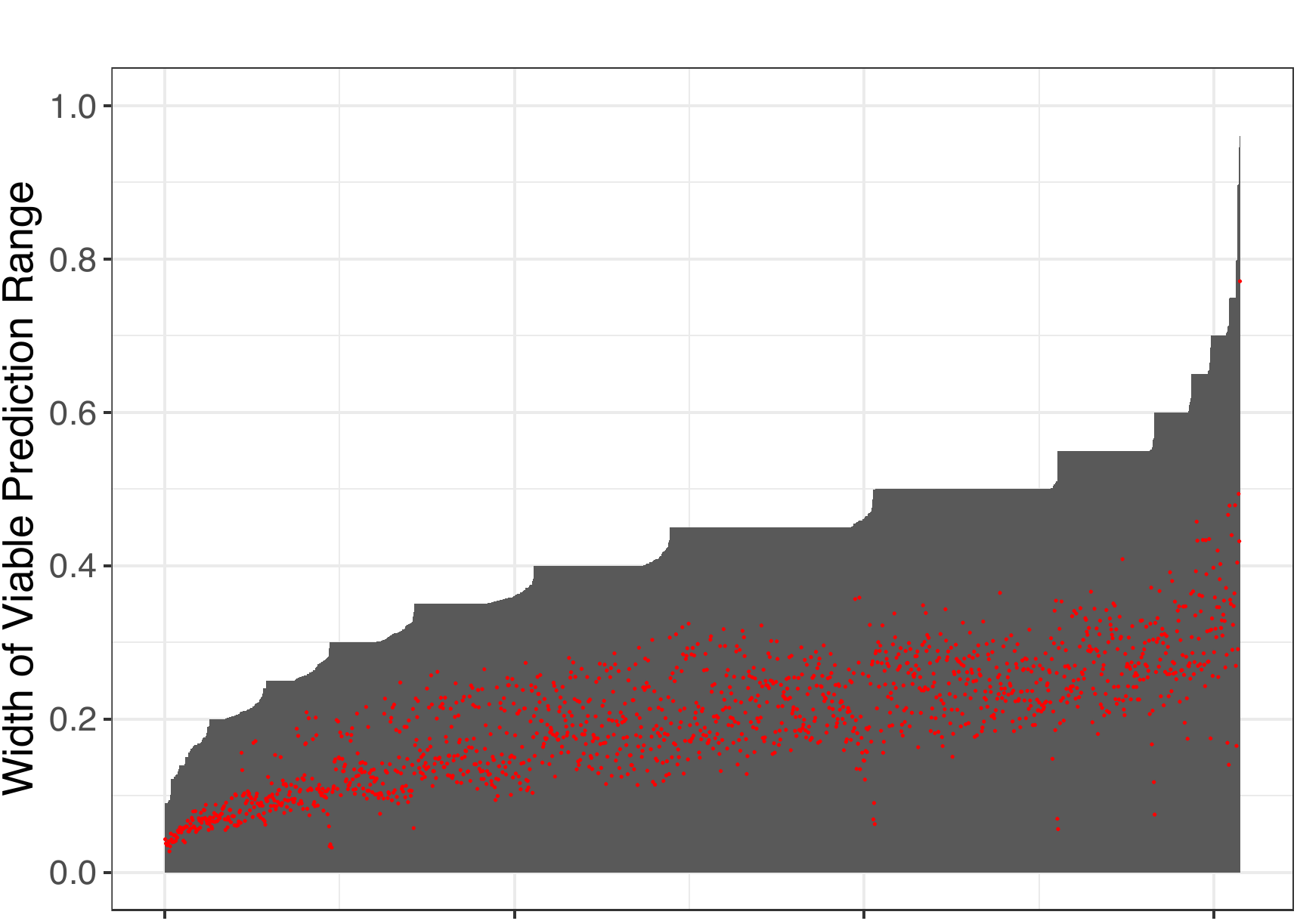} &
    \addinterval{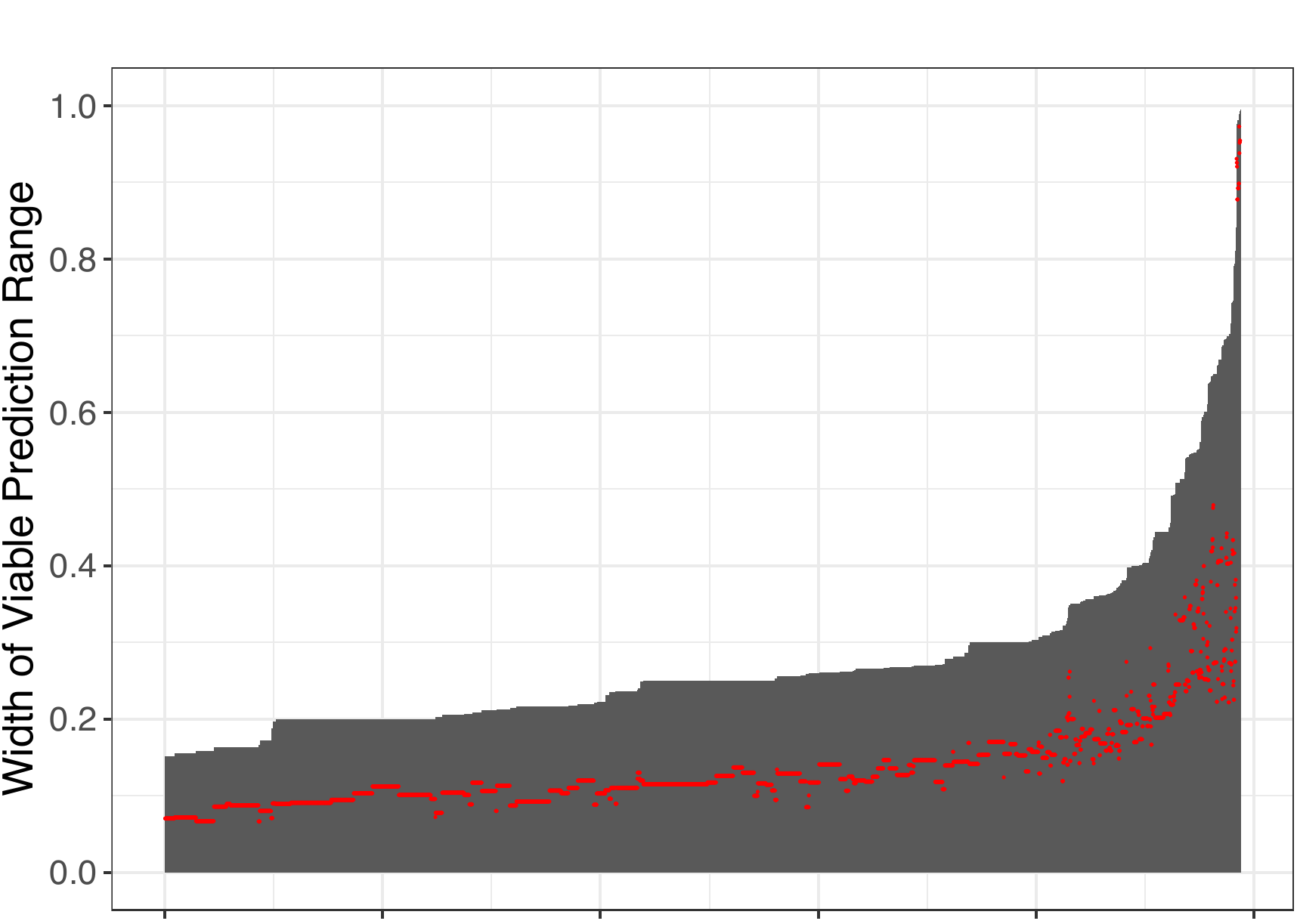} \\
    \midrule
    {\circled{B} } &
    \addinterval{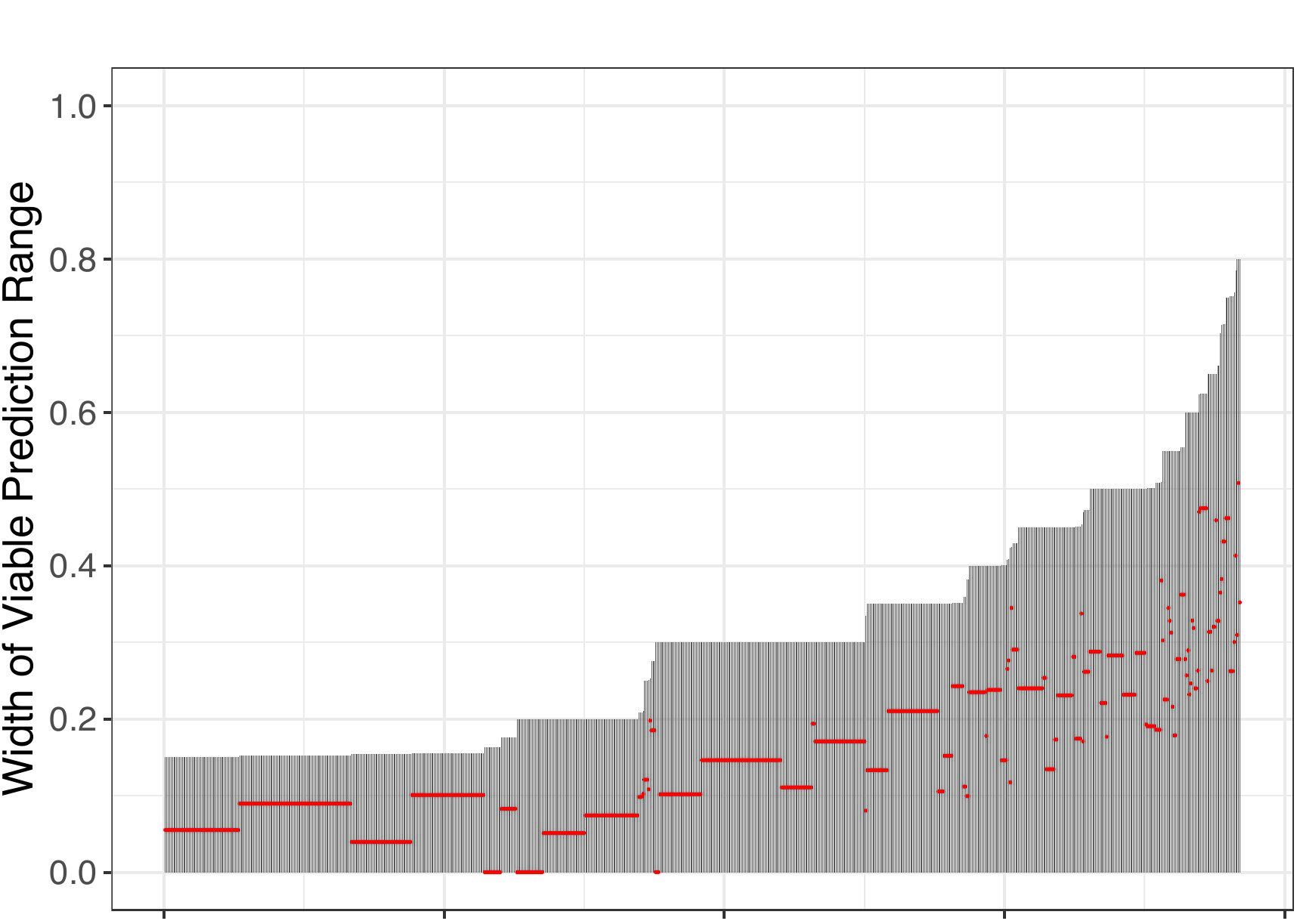} & 
    \addinterval{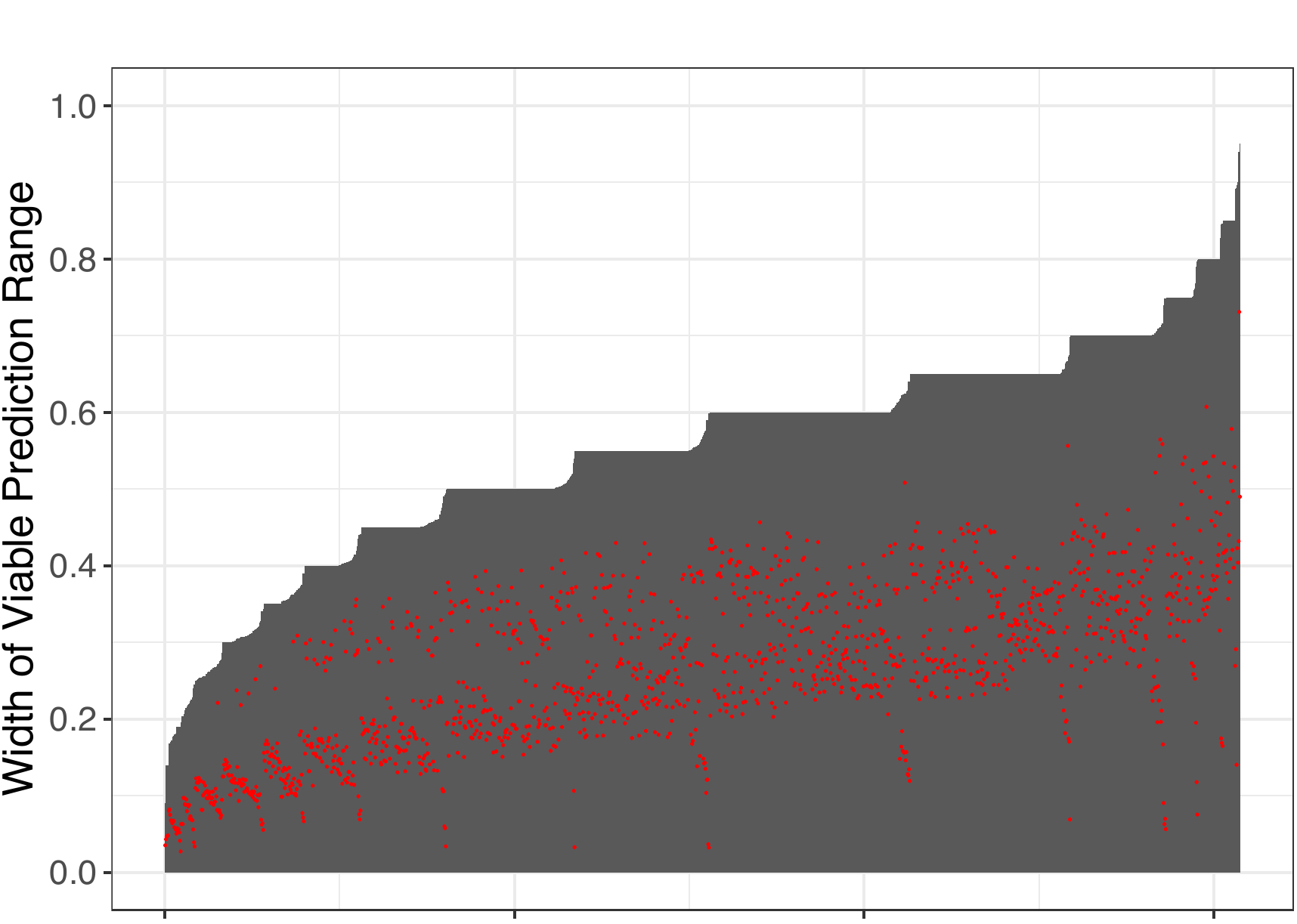} & 
    \addinterval{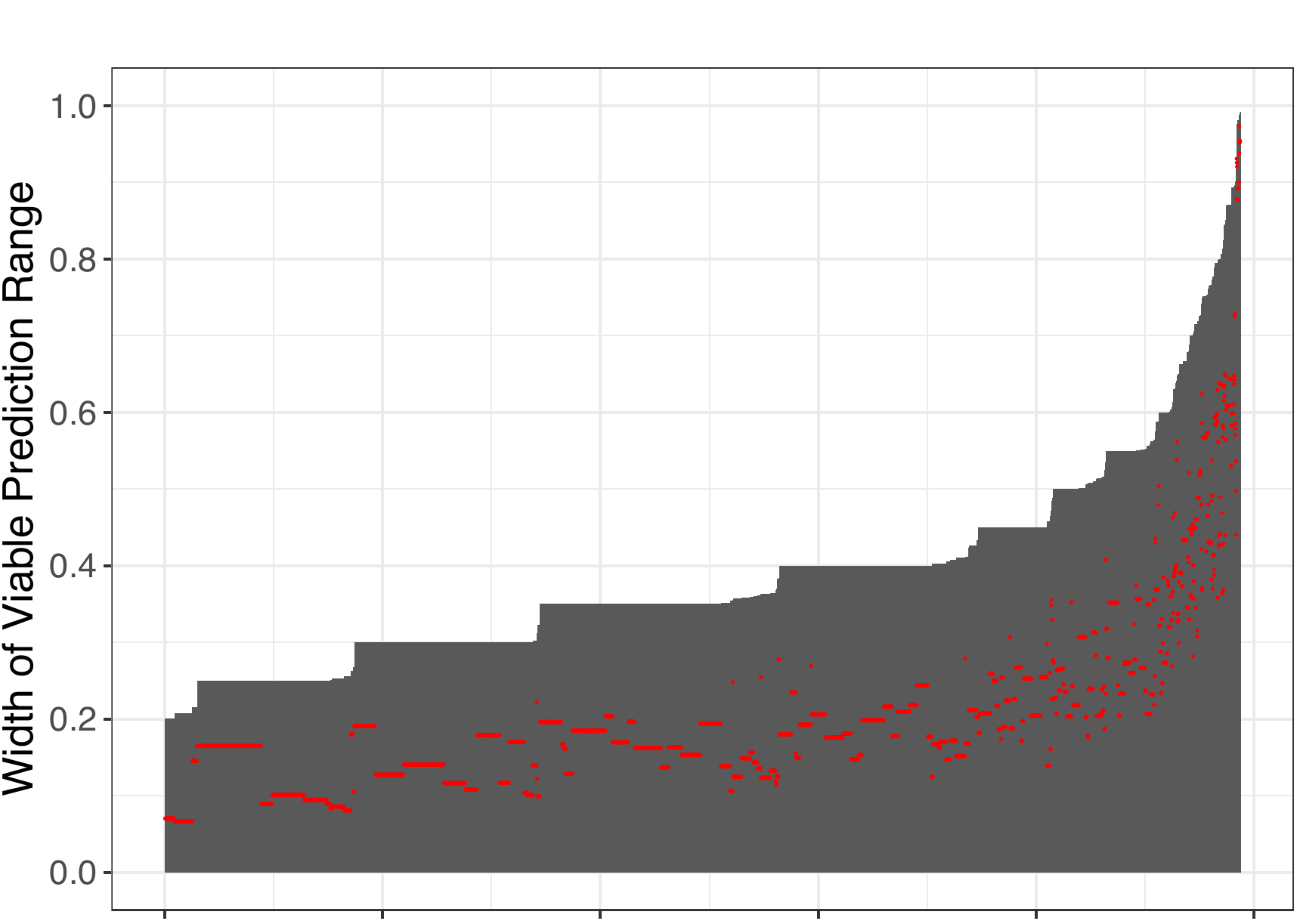} \\
    \midrule
    {\circled{C}} &
    \addhmFINAL{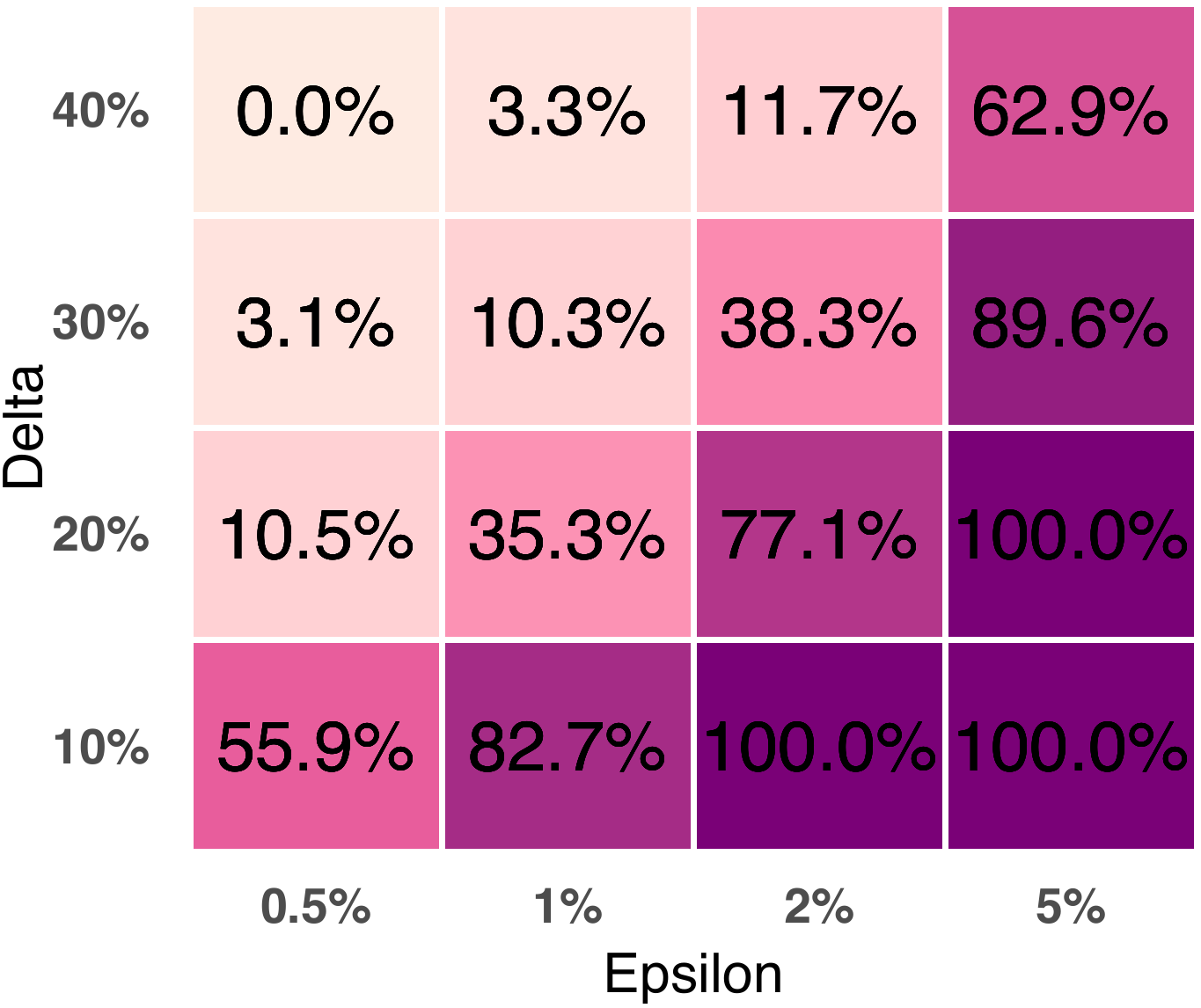} &
     \addhmFINAL{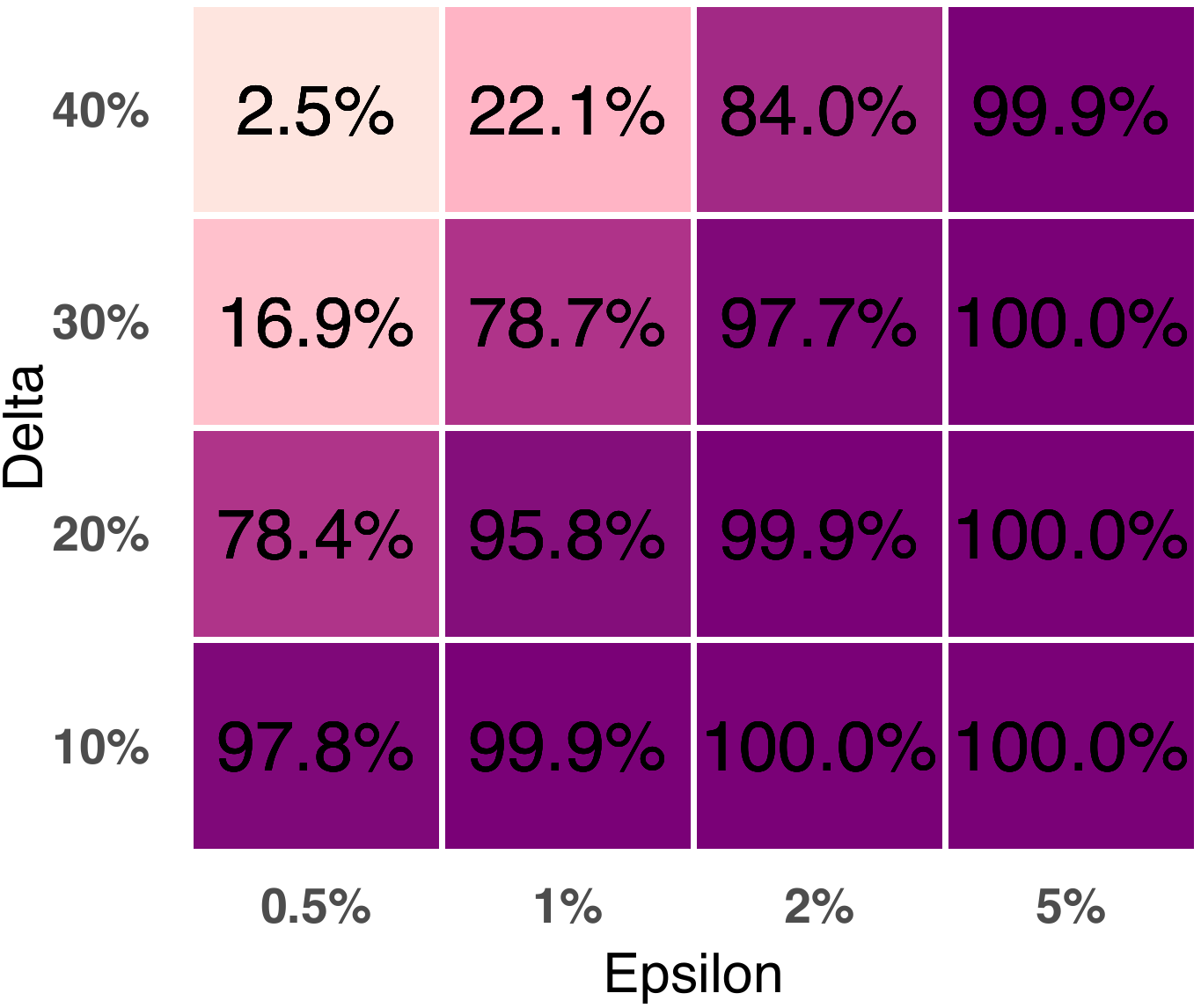} & 
    \addhmFINAL{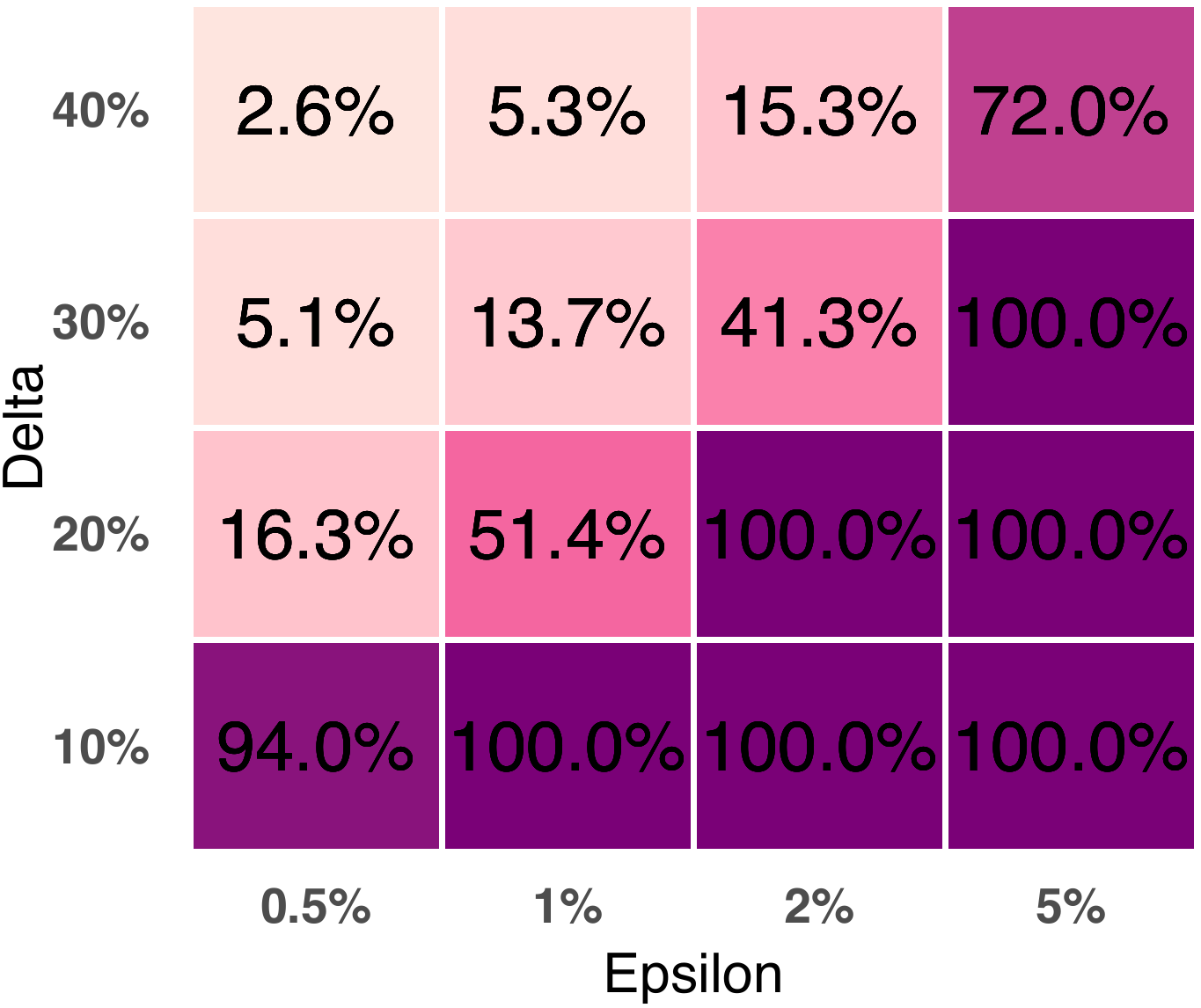} \\
    \midrule
    {\circled{D}  } &
    \addhmFINAL{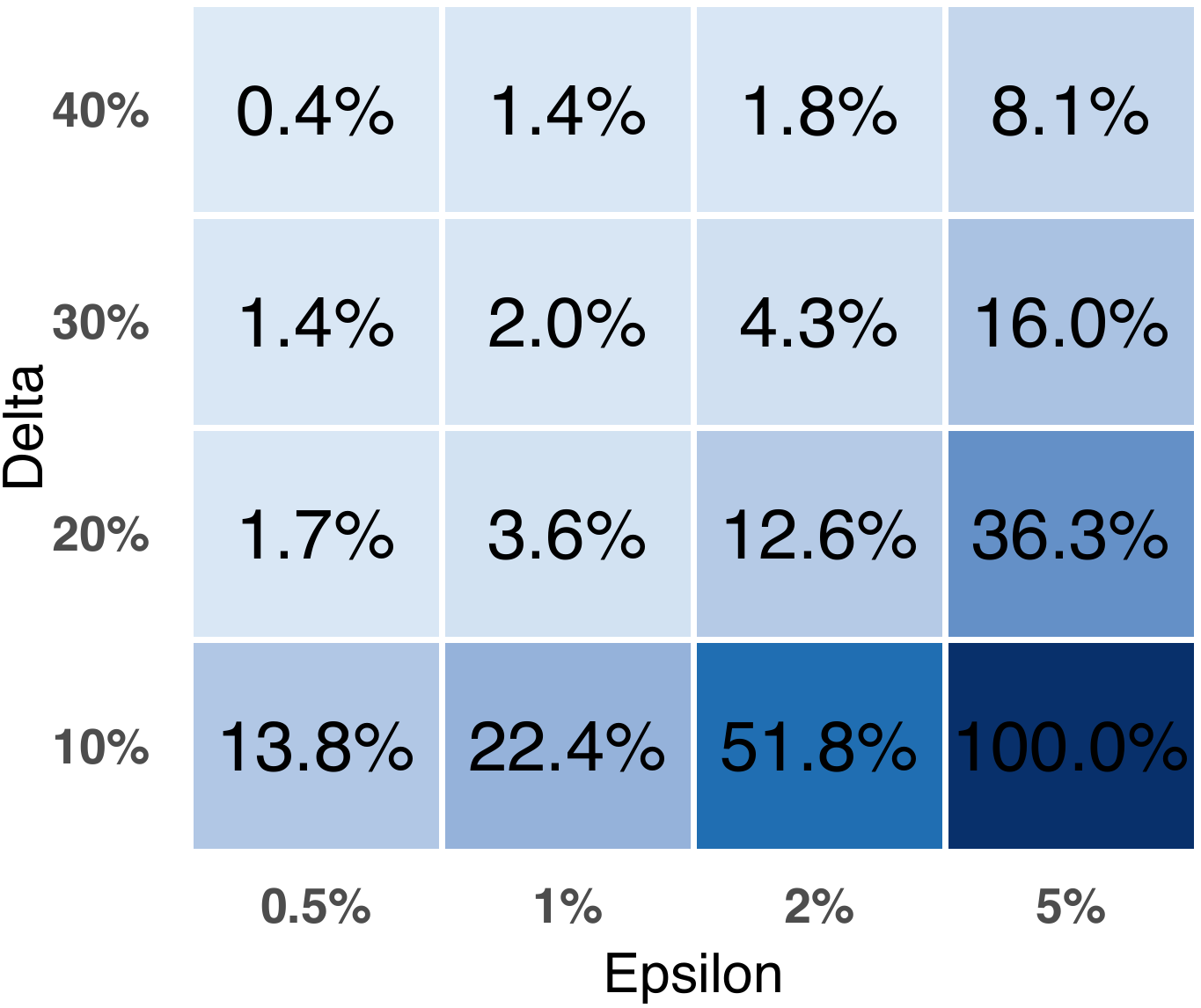} &
     \addhmFINAL{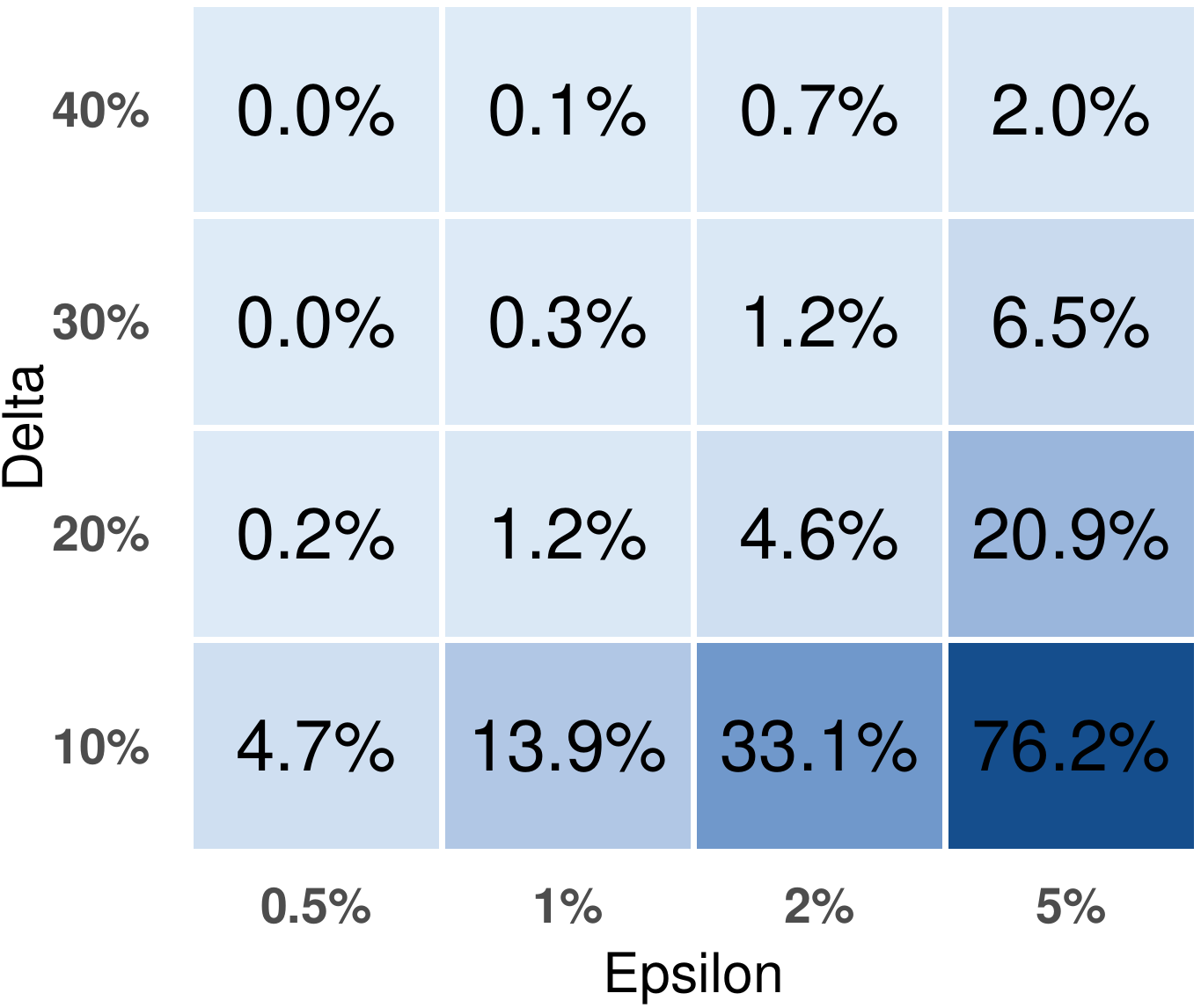} &
      \addhmFINAL{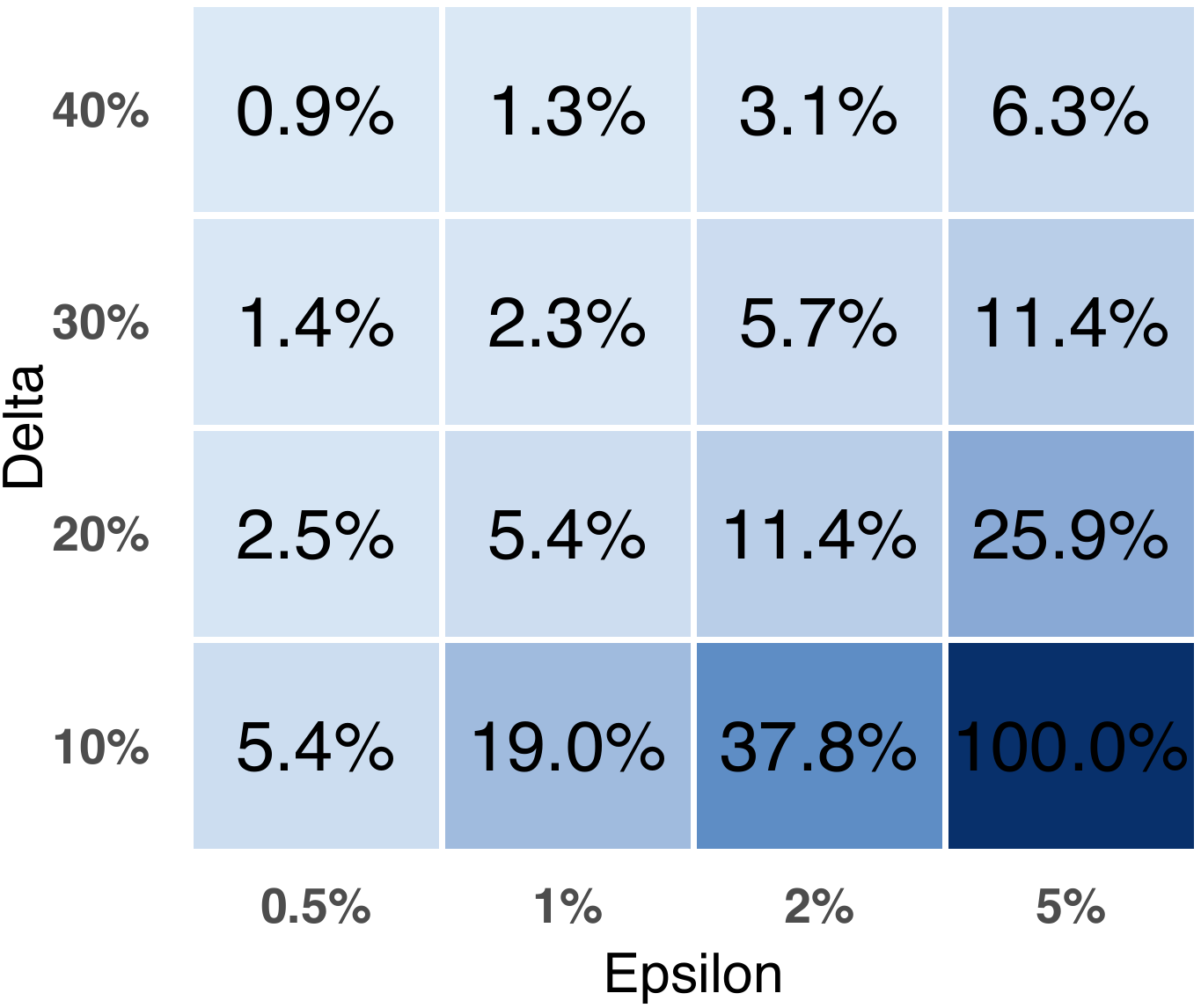} \\
    \bottomrule
    \end{tabular}}
    \caption{Predictive multiplicity in probabilistic classification on \textds{mammo}, \textds{apnea} and \textds{arrest}. In rows (A) and (B) we show the distribution of viable prediction ranges $|V_\epsilon(\xb_i)|$ on the y-axis for each example on the x-axis (relative baseline estimates in red). Notice, pointwise viable prediction ranges are plotted in increasing order from left to right. We plot viable prediction ranges for competing models with near-optimal training AUC~(A) and training loss~(B). See illustration in Figure~\ref{fig:range_to_amb}. We also show ambiguity~(C) and discrepancy~(D) for competing models with respect to training loss. We include other datasets in Appendix E.\label{Fig::HeatmapsAmbiguityDiscrepancyAll} }
\end{figure*}

\subsection{Real-World Datasets}
\label{Sec::Empirical}

In this section, we evaluate predictive multiplicity in risk prediction tasks from medicine, lending, and criminal justice.\footnote{This is not an endorsement of current usage of risk assessment tools in criminal justice. The use of prediction software raises serious concerns in this domain. We do not condone building models on arrest data to inform or justify increased policing.} Altogether, we consider seven datasets that exhibit variations in sample size, number of features, and class imbalance (see Table 1 in the Appendix). 
For each dataset, we compute viable prediction ranges, ambiguity and discrepancy using the methods outlined in \S\ref{Sec::Methodology}. When training candidate models, we adopt a grid of target predictions: $P = \{0.01, 0.1 , 0.2, \ldots, 0.9 , 0.99\}$. 
We  compute discrepancy by solving the MINLP Eq.~\eqref{Eq::DiscERM} with CPLEX v20.1~\citep{Diamond2016} on a single CPU with 16GB RAM. Our results are shown in Figure \ref{Fig::HeatmapsAmbiguityDiscrepancyAll}, and additional results are in the Appendix.

\paragraph{Viable Prediction Ranges.} Our results show that competing models can assign risk estimates that vary substantially. Viable prediction ranges are plotted in rows \textbf{(A)} and \textbf{(B)} of Figure~\ref{Fig::HeatmapsAmbiguityDiscrepancyAll}, and we see non-zero viable prediction ranges for all examples across all datasets. The viable ranges for \textds{apnea} and \textds{mammo} appear much larger compared to \textds{compas\_arrest}. In terms of near-optimal loss, \textds{apnea} has the most variation, while \textds{mammo} has the most variation in terms of AUC. This points to the value in varying near-optimal metric.

\paragraph{Ambiguity and Discrepancy.}

Ambiguity and discrepancy are shown in rows \textbf{(C)} and \textbf{(D)} of Figure~\ref{Fig::HeatmapsAmbiguityDiscrepancyAll}, respectively. For $\epsilon = 1\%$ and $\delta = 20\%$, we see ambiguity values at $35.3\%$ (\textds{mammo}), $95.8\%$ (\textds{apnea}), and $51.4\%$ (\textds{compas\_arrest}). This means that $35.3\%$ of breast cancer risk estimates vary by at least $20\%$ over near-optimal models. We see discrepancy values at $3.6\%$ (\textds{mammo}), $1.2\%$ (\textds{apnea}), and $5.4\%$ (\textds{compas\_arrest}) for $\epsilon = 1\%$ and $\delta = 20\%$. \textds{compas\_arrest} is the worst in terms of discrepancy, while \textds{apnea} has the most severe ambiguity. Thus, ambiguity and discrepancy are not always coupled.

\paragraph{On the Choice of Performance Metric.}
In settings where we want a model that performs well in terms of AUC, we should measure predictive multiplicity over a set of competing models with near-optimal AUC. In practice, it is often convenient to measure predictive multiplicity over a set of competing models that attain near-optimal loss (since the loss can be encoded into an optimization problem). This is a problem because small variations in loss can lead to large variations in AUC -- thus models with near-optimal loss may not match models with near-optimal AUC. Our results show that measures of predictive multiplicity vary considerably based on the performance metric used to define the set of competing models. In particular, we find that discrepancy and ambiguity will vary when measured over competing models that attain near-optimal loss, AUC, or ECE. 

\paragraph{On Samples Prone to Ambiguity.}
Our results reveal a relationship between ambiguity and individual {\em uniqueness} (number of duplicates), {\em class} imbalance, and {\em baseline risk estimate}. For uniqueness, we find that across datasets, less than $10\%$ of examples with more than $20$ duplicates are ambiguous. That unique examples are more prone to ambiguity is related to our findings on outliers (see \S\ref{Sec::ToyExamples}). %

In terms of class imbalance, we find datasets with class imbalance skewed negative (\textds{adult}, \textds{bank}) often exhibit multiplicity on positive examples. In comparison, datasets that are roughly balanced by class (e.g., \textds{mammo}, \textds{compas\_arrest}) have the same level of ambiguity for each class. This can be interpreted in light of the majority-minority effect from \S\ref{Sec::ToyExamples}.

In terms of the baseline risk estimate, we see high ambiguity for examples with baseline risk near 50\% on all datasets. For instance, all examples with baseline risk between $45\%$ and $55\%$ are ambiguous for the \textds{mammo} dataset  ($\epsilon = 0.5\%$ AUC, $\delta = 20\%$). There is no reason to believe that high ambiguity is less problematic for these samples. Rather, the importance of ambiguity will depend on the risk thresholds that drive decisions in a particular domain.

\paragraph{On the Disparate Impact of Multiplicity.}
Our results demonstrate how multiplicity can disproportionately impact individuals from historically marginalized groups. For example, when predicting the risk of rearrest, individuals who are ethnically Hispanic are disproportionately affected by predictive multiplicity: ambiguity is $39\%$ for African Americans and $49\%$ for Caucasians, compared to $98\%$ for Hispanics ($\epsilon = 1\%$ and $\delta = 20\%$). Hence, reporting predictive multiplicity for subgroups can reveal important fairness considerations when testing models deployed throughout society.

\section{Concluding Remarks}
\label{Sec::Conclusion}

We developed methods to evaluate the effect of slightly perturbing optimal model performance, revealing that similar models do not always assign similar predictions. We studied how competing models can assign conflicting predictions in probabilistic classification tasks. The proposed optimization-based methods compute our simple measures reliably. Compared to previous work, our methods allow for flexibility in choosing near-optimal metric and deviation threshold. Using synthetic data, we also present the first study providing insight into the kinds of data characteristics that give rise to predictive multiplicity and show that separability, outliers and majority-minority structure are informative. Empirically, we reveal concerning levels of predictive multiplicity in high-stakes domains.

More research is needed to examine predictive multiplicity for other loss functions and model classes (our methods immediately generalize to linear models with convex loss functions). Also, it will be important to study how to effectively communicate these effects to practitioners and decision makers. Also, when a practitioner encounters high predictive multiplicity, more work is needed on response options and mitigation strategies. Given predictive multiplicity metrics, practitioners can make better decisions in model selection while end-users can adjust their reliance on individual risk predictions. Concisely, analyzing predictive multiplicity promotes accountability and transparency in machine learning.

\section*{Acknowledgements}
We thank Yiling Chen, Ariel Procaccia, Elena Glassman, and Harvard EconCS group for feedback and helpful discussions. JWD is supported by a Ford Foundation Pre-doctoral Fellowship and the NSF  Graduate Research Fellowship Program under Grant No. DGE1745303. Any opinions, findings, and conclusions or recommendations expressed in this material are those of the author(s) and do not necessarily reflect the views of the NSF.
%

\fontsize{9.5pt}{10.5pt} 
\bibliography{references.bib}

\clearpage

\end{document}